\PassOptionsToPackage{numbers}{natbib}
\documentclass{article}

\usepackage[final,main]{neurips_2026}

\usepackage[utf8]{inputenc}
\usepackage[T1]{fontenc}
\usepackage{hyperref}
\usepackage{url}
\usepackage{booktabs}
\usepackage{amsfonts}
\usepackage{amssymb}
\usepackage{amsmath}
\usepackage{nicefrac}
\usepackage{microtype}
\usepackage{xcolor}
\usepackage{algorithm}
\usepackage{algpseudocode}
\usepackage{graphicx}
\usepackage{subcaption}
\usepackage{mathtools}
\usepackage{soul} 
\usepackage{xcolor}
\usepackage{caption}
\usepackage{placeins}

\newcommand{\Dcand}{\mathcal{D}_{\mathrm{cand}}}
\newcommand{\Dkin}{\mathcal{D}_{\mathrm{kin}}}

\newcommand{\Dfilt}{\mathcal{D}_{\mathrm{filt}}}

\newcommand{\E}{\mathbb{E}}
\newcommand{\bx}{\mathbf{x}}
\newcommand{\ba}{\mathbf{a}}
\newcommand{\bs}{\mathbf{s}}

\newcommand{\wangcan}[1]{#1}

\title{Reverse to Advance: Teleoperation-Cost Effective Hard Policy Learning from Reversed Easy Tasks}

\author{%
  Qiyuan Qiao\thanks{Equal contribution.}\quad
  Ge Yuan\footnotemark[1]\quad
  Can Wang\thanks{Corresponding author.}\quad
  Dong Xu\\
  The University of Hong Kong\\
  \textnormal{qiaoqy@connect.hku.hk}, \textnormal{gavinyuan97@gmail.com}\\
  \textnormal{\{canwang,dongxu\}@hku.hk}
}

\hypersetup{hidelinks, pdftitle={Reverse to Advance: Teleoperation-Cost Effective Hard Policy Learning from Reversed Easy Tasks}, pdfauthor={Qiyuan Qiao, Ge Yuan, Can Wang, Dong Xu}}

\begin{document}

\maketitle

\begin{abstract}
\wangcan{High-quality teleoperation datasets are costly to collect, particularly for hard tasks.
We observe that many tasks exhibit directional asymmetry: completing the forward hard task is difficult, whereas reversing it by relaxing or disrupting the environment is comparatively easy.
This suggests that reversed easy-task trajectories can serve as a scalable supervision signal for the hard task, reducing the cost of manual demonstration collection.
However, reversed data can be noisy, and directly training on it may yield suboptimal policies.
To enable largely automated acquisition and effective use of reversed data, we propose a teleoperation-cost effective framework for hard policy learning via temporal reversal of easy tasks, consisting of three key components: a closed-loop data collection pipeline that alternates between hard-task and easy-task policies to autonomously reset the environment and generate diverse trajectories; a hierarchical data refinement pipeline that temporally inverts easy-task rollouts and filters low-quality motion using kinematic priors and a critic-guided advantage filter; and an iterative policy learning method that trains the hard-task policy using both initial reversed easy-task demonstrations and the filtered reversed data in a continuous online learning loop. By combining automated collection, hierarchical refinement, and iterative learning, our method enables scalable, reliable training of complex, high-precision manipulation tasks. Across two simulated benchmarks and real-robot experiments, we demonstrate that our method improves hard-task success rates with higher data efficiency and more stable training compared to reversal-based and reinforcement-learning baselines, without requiring extensive hard-task teleoperation.
}

\end{abstract}

\section{Introduction}
\label{sec:introduction}

\wangcan{
Learning complex robotic manipulation tasks remains a fundamental challenge in embodied AI. 
While Imitation Learning~\cite{hussein2017imitation,zare2024survey}, particularly through Diffusion Policies~\cite{chi2023diffusion}, has shown remarkable prowess in mapping high-dimensional observations to precise actions, its performance is heavily bottlenecked by the availability and quality of expert demonstrations.
However, collecting sufficient demonstrations is an arduous process.
This process demands unwavering concentration from human operators, as even a split-second lapse in focus can result in a failed trial~\cite{gao2024efficient}.
Furthermore, the quality of human-generated data is inherently inconsistent; datasets are frequently plagued by suboptimal paths, unintended oscillations, and varying levels of expertise, which can introduce significant noise and hinder the policy's ability to converge on a robust solution.
These limitations are particularly exacerbated in hard tasks characterized by high-precision requirements and complex contact dynamics.

We observe that in many manipulation scenarios, there exists a fundamental asymmetry in task difficulty: solving a task, such as inserting a peg into a hole, stacking delicate blocks, or assembling a complex mechanism, is significantly harder than unsolving it, such as extraction, de-stacking, or disassembly. While the forward hard task often requires navigating a narrow state-space funnel with high-precision contact, the reverse easy task typically involves expanding into free space, making it far more amenable to both human teleoperation and automated execution.
This observation provides a key insight: \textit{Can we leverage the abundant, easily collected demonstrations from these reverse tasks to bootstrap the learning of the original hard task?} If a robot can master the easy reverse task, the resulting trajectories, when temporally inverted, could theoretically serve as a rich source of supervision for the hard task. This approach promises to drastically reduce the human labor required by substituting high-effort expert demonstrations with low-effort, scalable data from the task's inverse. 
However, simply reversing these trajectories is often insufficient, as they lack the corrective behaviors needed to handle the precision requirements and covariate shift inherent in the hard-task domain.
}

\wangcan{To the best of our knowledge, existing approaches rarely exploit this forward–reverse task asymmetry. Prior work either relies on direct Reinforcement Learning from scratch~\cite{jiangtime,cheng2023look,yao2023learning} or on standard data augmentation with the reversed data~\cite{barkley2024investigation}. However, these methods still fail to account for the inherent noise and suboptimal behaviors present in reversed trajectories.
These observations motivate the design of a framework for learning complex manipulation policies under teleoperation cost constraints via simple-task temporal reversal. To achieve this, several key challenges must be addressed: 1) \textbf{Automatic data collection:} How can we construct a largely automated pipeline for acquiring reversed-task demonstrations, thereby minimizing the human teleoperation effort? 2) \textbf{Data filtering: }How can we identify and extract the most valuable, high-quality trajectories from the noisy reversed data to serve as reliable supervision for the original hard task? 3) \textbf{Progressive learning:} Is it possible to integrate automated data collection and filtering into an online learning loop, where the policy continuously improves by leveraging filtered data, allowing the system to progressively approach optimal performance on the hard task?
}

\wangcan{
To address these challenges, we propose 
\textbf{Auto-E2H} (\textbf{Auto}mated \textbf{E}asy-\textbf{to}-\textbf{H}ard Data Acquisition and Refinement),
a teleoperation-cost effective framework for hard policy learning via temporal reversal. 
To efficiently collect high-quality data with minimal human intervention, we first design a \textbf{closed-loop data collection pipeline} that alternates between hard-task and easy-task policies. This mechanism enables the policies to reset each other, generating diverse trajectories while significantly reducing the need for teleoperation. 
To filter noise in the collected data, we further design a \textbf{hierarchical data refinement pipeline}, where easy-task rollouts are temporally reversed and processed through kinematic filtering to remove slow or static segments, followed by critic-guided advantage filtering to retain only high-value trajectory segments.
We further propose an \textbf{iterative policy learning method}, where the hard-task policy is trained using both the initial reversed easy-task demonstrations and the filtered reversed data, forming a continuous online learning loop that progressively improves performance.
By combining automated collection, hierarchical refinement, and iterative policy learning, our framework ensures that the hard-task policy is trained on informative, diverse, and reliable trajectories, enabling scalable learning of complex, high-precision manipulation tasks with minimal human supervision.
}


\section{Related Works}
\label{sec:related}

\wangcan{
\textbf{Imitation Learning in Robotics.} Imitation Learning~\cite{hussein2017imitation,zare2024survey} aims to train policies by leveraging demonstrations from expert behavior. Classic frameworks include Behavioral Cloning~\cite{florence2022implicit,torabi2018behavioral,foster2024behavior,mehta2025stable,sasaki2020behavioral,wang2025language,bai2025rethinking} and more recent methods based on generative modeling, such as diffusion policies~\cite{chi2023diffusion}, which have shown strong capabilities in mapping high-dimensional observations to precise actions. 
However, the success is heavily dependent on the quality and quantity of demonstrations. Complex and high-precision manipulation tasks often require substantial human effort to collect consistent, expert-level trajectories, and suboptimal or noisy demonstrations can significantly hinder policy performance~\cite{chen2025restoring}.
To address this challenge, we propose a teleoperation-efficient framework that uses easily collected reversed-task trajectories to aid policy learning of the corresponding forward, complex tasks, reducing reliance on large-scale high-quality expert data.

\textbf{Data Collection and Refinement in Teleoperation.} Collecting high-quality demonstrations remains a major bottleneck for imitation learning. Recent work has focused on improving efficiency and data utility. Some approaches use portable or universal teleoperation systems to capture rich, multi-task demonstrations across robots and environments, reducing repeated human effort~\cite{chi2024universal,wang2024dexcap}. Others enhance learning from limited data via augmentation, human-in-the-loop refinement, or pre-trained multimodal models~\cite{luo2025precise,shukor2025smolvla,wen2025tinyvla,bharadhwaj2024roboagent}. Compositional and structured strategies exploit environment variations and scaling laws to improve generalization and maximize the value of each demonstration~\cite{gao2024efficient,lin2024data}.
Recent works have considered how to refine the collected data.
For example, GR-RL~\cite{li2025gr} adopts a learned task progress model to identify low-quality transitions in human demonstrations, discarding segments that do not contribute positively to task progress before applying offline and online reinforcement learning. 
TR-DRL~\cite{jiangtime} exploits dynamics-consistent filters to identify fully reversible transitions, augment trajectories by reversing them, and apply potential-based reward shaping to partially reversible components. 
In contrast, we propose a hierarchical data refinement method that combines kinematic filtering and critic-guided advantage evaluation. Unlike prior work, our approach removes low-quality or static segments and retains only those contributing positively to task success, enabling more effective use of reversed-task trajectories.

\paragraph{Forward–Reverse Task Asymmetry.} 
In many manipulation scenarios, there exists a natural asymmetry in task difficulty: the forward task often involves precise, constrained interactions, whereas the corresponding reversed task is typically easier. Leveraging this asymmetry has the potential to reduce human teleoperation effort by using easily collected reversed-task trajectories to supervise learning of the harder forward tasks.
Existing works have explored temporal or time-reversal symmetries to exploit this structure in RL~\cite{goyalrecall,yao2023learning,cheng2023look,barkley2024investigation,jiangtime} and self-learning~\cite{nair2020trass}. TR-DRL~\cite{jiangtime}, for example, identifies fully reversible transitions using a dynamics-consistent filter, augments these transitions by reversal, and applies potential-based reward shaping to partially reversible components, thereby improving sample efficiency in both single- and multi-task RL benchmarks. However, such methods primarily focus on enforcing physical reversibility or shaping rewards based on object states, and they do not explicitly address the combination of data collection, noise filtering, and policy learning from reversed trajectories.
In contrast, our framework integrates automated collection, hierarchical refinement, and iterative policy learning to leverage reversed-task data effectively, enabling scalable learning of complex forward tasks with minimal human intervention.
}




\section{Method}
\label{sec:method}

\wangcan{\textbf{Problem Formulation.}} \wangcan{
Consider a finite-horizon manipulation task of interest, denoted by $\mathcal{T}_h$, which exhibits a high degree of difficulty due to complex dynamics. We define its corresponding \emph{reverse-direction task} $\mathcal{T}_e$, which is assumed to be easier to solve. Both tasks share a common observation space $\mathcal{S}$, action space $\mathcal{A}$, and horizon $H$. 
Formally, let $\tau_e=\left \{ (\mathbf{s}_t, \mathbf{a}_t) \right \}_{t=0}^{H-1}$ denote a successful trajectory of $\mathcal{T}_e$. We assume the transformed trajectory $\tilde{\tau}_h =\mathrm{Reverse}(\tau_e)$ contains state-action segments 
that provide meaningful supervision for learning a policy on $\mathcal{T}_h$. 
Our goal is to learn a policy $\pi_h$ for the hard task $\mathcal{T}_h$ using a limited initial hard-direction dataset $\mathcal{D}_h$ obtained by reversing easy-task demonstrations, together with additional easy-task data $\mathcal{D}_e$:
\begin{equation}
    \pi_h^* = \arg\max_{\pi_h}\mathbb{E}_{\tau_h\in\mathcal{D}_h, \tau_e\in\mathcal{D}_e} \Big[ R(\tau_h, \tilde{\tau}_h) \Big| \tilde{\tau}_h = \mathrm{Reverse}(\tau_e) \Big].
    \label{eq:1}
\end{equation}
where $R(\cdot)$ denotes the success rate of the deployment of the hard task $\mathcal{T}_h$.
}

\begin{figure}[t]
    \centering
\includegraphics[width=1.0\linewidth]{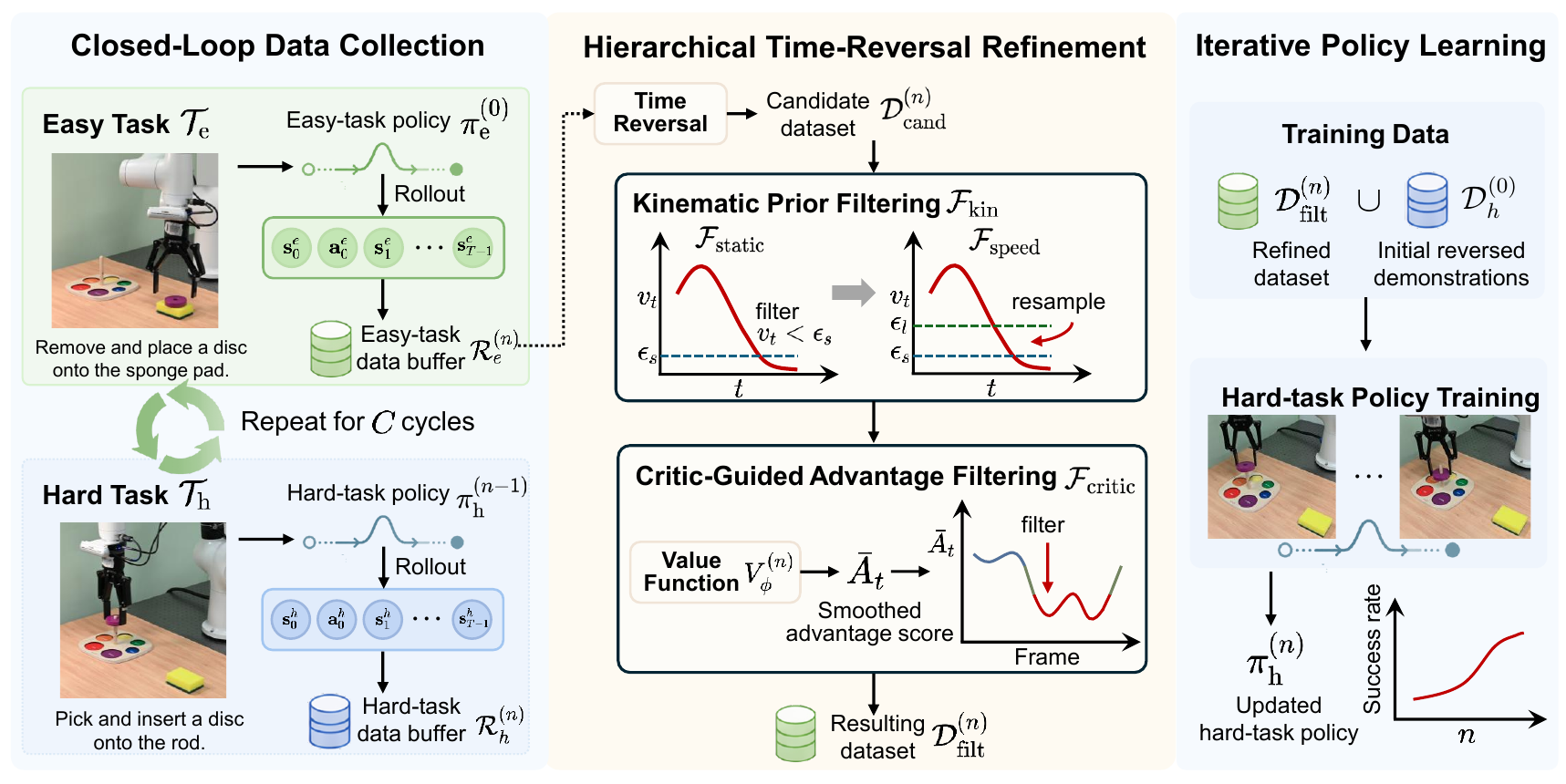}
    \caption{\textbf{Framework.} 
    \wangcan{Our framework consists of three key components. 
    First, the Closed-Loop Data Collection stage alternates between executing the easy-task policy $\pi_e^{(0)}$ and the evolving hard-task policy $\pi_h^{(n-1)}$ over multiple cycles, generating diverse trajectory data buffers for each task. In the Hierarchical Time-Reversal Refinement step, easy-task trajectories are reversed and undergo kinematic filtering to remove slow or static segments, followed by critic-guided advantage filtering to prune low-progress segments. Finally, Iterative Policy Learning updates $\pi_h^{(n)}$ using the fixed initial dataset $\mathcal{D}_h^{(0)}$ and the current refined dataset $\mathcal{D}_{\mathrm{filt}}^{(n)}$.}
    }
\label{fig:pipeline}
\end{figure}

\wangcan{
\textbf{Overview.}
To address this challenge, we introduce
\textbf{Auto-E2H},
a time-reversal-based self-improving framework for hard-task policy learning, as shown in Figure~\ref{fig:pipeline}.
First, to enable efficient, scalable data collection and support online self-improvement, we expect to execute the hard-task and easy-task policies in an alternating closed loop, allowing each policy to reset the environment for the other and alleviating repeated manual interventions. 
To implement this, we propose a closed-loop demonstration collection method (\textbf{Section 3.1}) that ensures continuous and diverse trajectory generation for both tasks. 
Second, the reversed trajectory $\mathrm{Reverse}(\tau_e)$ can be very noisy for learning the hard-task policy $\pi_h$. Thus, we propose a hierarchical refinement process (\textbf{Section 3.2}) to obtain high-quality candidate segments, ensuring that the trajectories $\tilde{\tau}_h = \mathrm{Reverse}(\tau_e)$ provide meaningful supervision for hard-task policy learning.
Third, to iteratively improve the hard-task policy $\pi_h$, we propose an iterative policy learning algorithm (\textbf{Section 3.3}) that updates $\pi_h$ using both the initial reversed easy-task demonstrations and the refined reversed trajectories.
}

\subsection{Closed-Loop Data Collection}
\label{sec:closed_loop}

\wangcan{
Since our goal is to enable online self-improvement of the policy, data collection must be efficient, scalable, and minimally reliant on human intervention. 
By leveraging replay rollouts from the action policy, Auto-E2H substantially reduces the cost of human teleoperation, while the stochastic denoising process inherent to diffusion policies further promotes behavioral diversity in the collected data. 
However, a key challenge is the need for frequent environment resets: after executing the hard task, the system must return to a valid initial configuration for the next trial. 
Auto-E2H resolves this issue through an autonomous closed-loop collection strategy, where the hard-task and easy-task policies are executed in an alternating manner to reset the environment for each other, alleviating the need for repeated human intervention. 

Specifically, at training iteration $n$, we perform $C$ collection cycles. In each cycle $c \in \{1,\dots,C\}$, the process consists of two stages: 1) \textbf{Hard-task rollout}: we execute the current hard-task policy $\pi_h^{(n-1)}$ on $\mathcal{T}_h$ for up to $H$ steps, yielding a trajectory $\tau_h^{(c)}$; 2) \textbf{Easy-task rollout}: starting from the terminal state of $\tau_h^{(c)}$, we execute the fixed easy-task policy $\pi_e^{(0)}$ on $\mathcal{T}_e$ for up to $H$ steps, yielding a trajectory $\tau_e^{(c)}$.
The terminal state of $\tau_e^{(c)}$ then serves as the initial state for the next hard-task rollout, forming a closed interaction loop. After an external reset, this process can repeat without manual intervention while automatic resets succeed. If a reset fails, collection pauses, the arm returns home, and an operator restores the scene before collection resumes.
We index the hard-task and easy-task trajectory buffers by the collection iteration $n$:
\begin{equation}
    \mathcal{R}_h^{(n)} = \{\tau_h^{(c)}\}_{c=1}^{C},
    \quad
    \mathcal{R}_e^{(n)} = \{\tau_e^{(c)}\}_{c=1}^{C}.
    \label{eq:2}
\end{equation}
All trajectories in $\mathcal{R}_h^{(n)}$, including failures, are used for critic training. Only successful trajectories from $\mathcal{R}_e^{(n)}$ are used to construct candidate reversed data for policy learning.

This closed-loop mechanism expands the initial-state distribution for the hard-task policy, as the easy-task policy produces diverse terminal states. After reversal and transformation, these states yield varied candidate starting configurations for the hard task. While this diversity is beneficial, it also introduces noise, making a refinement stage necessary to filter out low-quality segments.
}

\subsection{Hierarchical Data Refinement}

\wangcan{
To optimize the collected data, we perform a three-step hierarchical refinement: data preprocessing, which reverses the trajectory; kinematic filtering, which removes slow or static segments and regularizes motion; and critic-guided advantage filtering, which prunes low-progress segments to align with the hard-task goal.
}

\wangcan{
\textbf{Data Preprocessing. }
For each successful easy-task rollout $\tau_e = \{(\bs_t, \ba_t)\}_{t=0}^{T-1}$ in the buffer $\mathcal{R}_e^{(n)}$, we construct a candidate hard-task trajectory by reversing the state sequence:
\begin{equation}
    \tilde{\tau}_h = \{\bs_{T-1-t}\}_{t=0}^{T-1}.
    \label{eq:3}
\end{equation}

Next, the action sequence is reconstructed from adjacent reversed end-effector poses and the gripper state rather than directly reusing the original actions: 
\begin{equation}
\ba'_t = h(\bx_t, \bx_{t+1}, g_t), \quad t = 0, \dots, T-2,
\label{eq:4}
\end{equation}
where $\bx_t=[x,y,z,q_w,q_x,q_y,q_z]$ is the reversed end-effector pose in the robot frame and $g_t\in\{0,1\}$ is the reversed gripper signal ($1$: open; $0$: closed). The function $h(\cdot)$ maps adjacent reversed poses and the gripper signal to the policy's 8-D absolute-pose and gripper-command representation. The last frame is discarded because it has no next-pose target. Directly reversing the original actions misaligns the pose targets, while negating absolute positions or quaternions does not invert motion. This reconstructs trajectory geometry rather than contact forces.
After these processing steps, we obtain the candidate dataset $\Dcand^{(n)}$ for the training iteration $n$:
\begin{equation}
    \Dcand^{(n)} = \left\{(\tilde{\tau}_h^{(c)},\ba'^{(c)}) \;\middle|\; 1\leq c\leq C,\ \tau_e^{(c)}\text{ succeeds}\right\}.
    \label{eq:5}
\end{equation}
The reversed trajectories in $\Dcand^{(n)}$ are diverse but noisy, and training directly on this data would degrade policy quality. Therefore, we apply Kinematic Prior Filtering followed by Critic-Guided Advantage Filtering:
\begin{equation}
    \Dfilt^{(n)} = \mathcal{F}_{\mathrm{critic}}\bigl(\mathcal{F}_{\mathrm{kin}}(\Dcand^{(n)})\bigr).
    \label{eq:6}
\end{equation}

\textbf{Kinematic Prior Filtering.} The kinematic filter is defined as: 
\begin{equation}
    \mathcal{F}_{\mathrm{kin}} = \mathcal{F}_{\mathrm{speed}} \circ \mathcal{F}_{\mathrm{static}}.
    \label{eq:7}
\end{equation}
The purpose of the static filter $\mathcal{F}_{\mathrm{static}}$ is to address segments of the trajectory where the motion is unnaturally slow. 
Such segments are statistically distinguishable through near-static or abnormally low-velocity kinematic profiles caused by inference latency, communication delay, or ambiguity-induced oscillations during action-policy rollouts. 
To handle this, we first calculate the instantaneous displacement $v_t$ of the end-effector between time steps $t$ and $t+1$:
\begin{equation}
    v_t = \|\bx_{t+1} - \bx_t\|_2.
    \label{eq:8}
\end{equation}
We then eliminate contiguous intervals where the frame-to-frame displacement remains below a threshold $\epsilon_s$, and the total spatial drift does not exceed $\epsilon_d$ over more than $L_{\mathrm{s}}$ consecutive steps. 
This effectively removes hovering and frozen segments that contribute little to no supervisory signal. 
The speed filter $\mathcal{F}_{\mathrm{speed}}$ targets segments whose velocity lies between the static threshold $\epsilon_s$ and a low-speed threshold $\epsilon_l$. For these segments, where the velocity is higher than $\epsilon_s$ but still low enough to warrant adjustment, we resample the motion to accelerate it toward a target step size $\Delta H$. This acceleration regularizes the action distribution, helping to ensure a more consistent and dynamic trajectory. 
Frames where the velocity exceeds the low-speed threshold $\epsilon_l$ are left unaltered, as they are considered to have sufficiently natural motion speeds.
After these processing steps, we obtain a kinematically cleaned dataset $\Dkin^{(n)}$.
}

\wangcan{
\textbf{Critic-Guided Advantage Filtering. }
Kinematic cleanup removes obviously degenerate frames but does not certify that the remaining segments make actual progress on the hard task.
We address this with Critic-Guided Advantage Filtering $\mathcal{F}_{\mathrm{critic}}$, where a critic trained on the \emph{current} hard-task rollouts is used to prune disadvantage segments from the kinematically cleaned dataset $\Dkin^{(n)}$.

At each iteration $n$, we train a value function $V_\phi^{(n)}(\bs)$ from scratch on all hard-task rollouts in $\mathcal{R}_h^{(n)}$. With success indicator $y\in\{0,1\}$, episode length $T\leq H$, per-step reward $-1$, terminal success reward $H$, and discount $\gamma{=}1$, the raw return is $G_t^{\mathrm{raw}}=-(T-t)+Hy$. We normalize it as $\bar{G}_t=(G_t^{\mathrm{raw}}-H)/(2H)$ and minimize
\begin{equation}
    \mathcal{L}_{\mathrm{critic}}(\phi) = \E_{\tau \sim \mathcal{R}_h^{(n)},\, t} \left[\bigl(V_\phi(\bs_t) - \bar{G}_t\bigr)^2 \right],
    \qquad
    \bar{G}_t = -\frac{T-t}{2H}-\frac{1-y}{2}.
    \label{eq:9}
\end{equation}
Even when all rollouts fail with $T=H$, $\bar{G}_t=-1+t/(2H)$ remains nonconstant. At cold start this provides temporal-ordering supervision, rather than a calibrated success probability.
We retrain the critic from scratch at every iteration so that it tracks the state distribution and failure modes of the current hard-task policy rather than a stale mixture of earlier policies.

For each reversed trajectory, we optionally truncate the suffix after it enters a high-value region to suppress irrelevant post-contact retreat phases. Then, with $\bar{r}=-1/(2H)$, we compute a per-step progress score using temporal-difference advantages:
\begin{equation}
    \delta_t = \bar{r} + \gamma V_{t+1} - V_t,
    \quad
    A_t = \sum_{l=0}^{T-t-1} (\gamma \lambda)^l \delta_{t+l}.
    \label{eq:10}
\end{equation}
Frames with the smoothed advantage score $\bar{A}_t < \alpha_{\mathrm{drop}}$ are marked for removal, and we ensure minimum run lengths for both retained and dropped segments to avoid fragmentation. The resulting dataset is $\Dfilt^{(n)}$.
This approach ensures that the critic only vetoes segments that do not align with the task's progress, reducing the risk of critic errors impacting the training process.
}

\subsection{Iterative Policy Learning}

\begin{algorithm}[ht]
\caption{Iterative policy learning with alternating hard-task and easy-task rollouts. The easy-task policy remains fixed; each hard-task policy update uses the initial demonstrations and current filtered data.}
\label{alg:iterative_policy_learning}
\small
\begin{algorithmic}[1]
\Require Easy-task demonstrations $\mathcal{D}_e^{(0)}$ and their reversed hard-direction dataset $\mathcal{D}_h^{(0)}$; iterations $N$; cycles $C$
\State Train $\pi_h^{(0)} \leftarrow \text{TrainPolicy}(\mathcal{D}_h^{(0)})$ and $\pi_e^{(0)} \leftarrow \text{TrainPolicy}(\mathcal{D}_e^{(0)})$
\For{$n = 1, 2, \ldots, N$}
    \State Collect hard-task and easy-task rollouts $\mathcal{R}_h^{(n)}$ and $\mathcal{R}_e^{(n)}$ using $\pi_h^{(n-1)}$ and $\pi_e^{(0)}$ (Eq.~\ref{eq:2})
    \State Invert and align successful rollouts from $\mathcal{R}_e^{(n)}$ to form $\mathcal{D}_{\mathrm{cand}}^{(n)}$ (Eq.~\ref{eq:5})
    \State Apply kinematic refinement: $\mathcal{D}_{\mathrm{kin}}^{(n)} \leftarrow \mathcal{F}_{\mathrm{kin}}(\mathcal{D}_{\mathrm{cand}}^{(n)})$ (Eq.~\ref{eq:7})
    \State Train value function $V_\phi^{(n)}$ from scratch on all hard-task rollouts $\mathcal{R}_h^{(n)}$ (Eq.~\ref{eq:9})
    \State Filter with the critic: $\mathcal{D}_{\mathrm{filt}}^{(n)} \leftarrow \mathcal{F}_{\mathrm{critic}}(\mathcal{D}_{\mathrm{kin}}^{(n)} \mid V_\phi^{(n)})$ (Eq.~\ref{eq:10})
    \State Update $\pi_h^{(n)} \leftarrow \text{TrainPolicy}(\mathcal{D}_h^{(0)} \cup \mathcal{D}_{\mathrm{filt}}^{(n)}; \pi_h^{(n-1)})$
\EndFor
\State \Return $\pi_h^{(N)}$
\end{algorithmic}
\end{algorithm}

\wangcan{We further propose an Iterative Policy Learning algorithm to progressively improve the hard-task policy using initial reversed easy-task demonstrations and refined easy-task rollouts (Algorithm~\ref{alg:iterative_policy_learning}). Initially, the policies are trained on $\mathcal{D}_h^{(0)}$ and $\mathcal{D}_e^{(0)}$. At iteration $n$, policies $\pi_h^{(n-1)}$ and $\pi_e^{(0)}$ collect buffers $\mathcal{R}_h^{(n)}$ and $\mathcal{R}_e^{(n)}$. Successful easy-task trajectories are inverted and aligned to form $\mathcal{D}_{\mathrm{cand}}^{(n)}$ (Equation~\ref{eq:5}), then refined by $\mathcal{F}_{\mathrm{kin}}$ (Equation~\ref{eq:7}).

The critic $V_\phi^{(n)}$ is trained on all current hard-task rollouts $\mathcal{R}_h^{(n)}$, including failures, using Equation~\ref{eq:9}. Critic-guided advantage filtering then produces $\mathcal{D}_{\mathrm{filt}}^{(n)}$ using Equation~\ref{eq:10}. We warm-start $\pi_h^{(n)}$ from $\pi_h^{(n-1)}$ and rebuild its training set at each iteration as $\mathcal{D}_h^{(0)}\cup\mathcal{D}_{\mathrm{filt}}^{(n)}$: the fixed initial reversed demonstrations and the current iteration's accepted data. The easy-task policy $\pi_e^{(0)}$ remains fixed throughout.
}

\section{Experiments}
\label{sec:experiments}

\begin{figure}[t]
    \centering
    \includegraphics[width=0.98\linewidth]{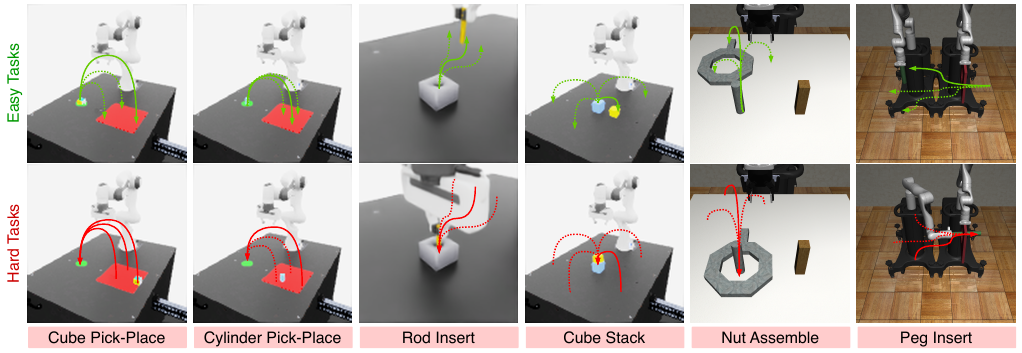}
    \caption{\wangcan{\textbf{Simulation benchmark task setups.}
    The first four pairs are Isaac Lab tasks spanning object relocation, contact-rich insertion, and multi-object stacking
    The last two pairs are Robosuite nut assembly/disassembly and peg insertion/removal.
    Within each pair, Task~$\mathcal{H}$ and Task~$\mathcal{E}$ share the same robot, observations, and action space but have opposite success directions.}}
    \label{fig:benchmark_demo}
\end{figure}
\begin{figure}[t]
    \centering
    \includegraphics[width=0.98\linewidth]{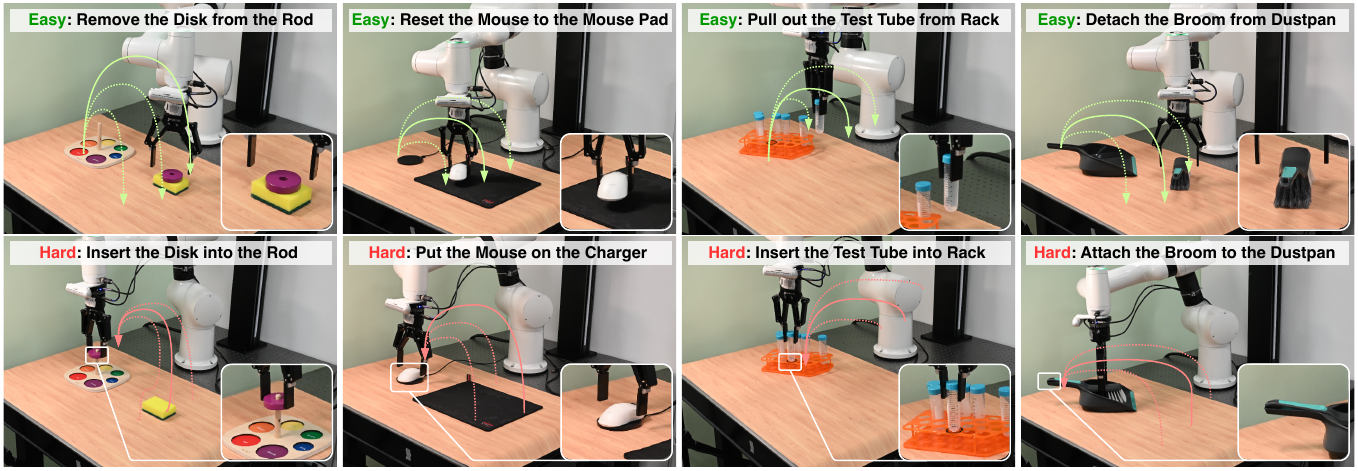}
    \caption{\textbf{Real-world robot task setups.}
    The four panels show near-completion photographs of the Disk, Mouse, Test Tube, and Brush tasks used in our real-world evaluation.
    In each task pair, Task~$\mathcal{H}$ and Task~$\mathcal{E}$ share a geometric path; their contact dynamics need not be exactly reversible.}
    \label{fig:realworld_demo}
\end{figure}

\textbf{Experimental Setup.} \wangcan{We evaluate our method on four Isaac Lab, two Robosuite\cite{robosuite2020}, and four real-world task pairs, each consisting of a target hard task Task~$\mathcal{H}$ and an easier reverse-direction task Task~$\mathcal{E}$, as shown in Figure~\ref{fig:benchmark_demo} and Figure~\ref{fig:realworld_demo}.
}
We report hard-task success rates over 40 evaluation trials per Isaac Lab task, 50 per Robosuite task, and 20 per real-world task.
We use a single set of filtering thresholds across simulation and real-world environments, without per-task tuning.

\wangcan{We compare our method with two RL-based approaches, TR-DRL~\cite{jiangtime}, which combines temporal reversal with RL for policy learning, and RECAP~\cite{intelligence2025pi}, an offline RL algorithm that learns from the rollouts of a hard policy, as well as a direct hard-demo baseline Direct-$\mathcal{H}$. 
To ensure a fairer comparison, we also compare with TR-DRL-DP, which uses dynamics-based reversal within the same Diffusion Policy collection and imitation-learning loop as Auto-E2H.}
For the Diffusion Policy comparisons using reversed easy-task data, baselines share the same initial demonstrations, backbone, and hyperparameters as Auto-E2H. The hard-demo BC comparison instead uses 50 directly collected hard-task demonstrations; the Robosuite TR-DRL comparison uses its original SAC implementation.
RECAP differs only by adding a one-dimensional indicator to the Diffusion Policy conditioning input to identify rollout source.

\subsection{Main Results}

\begin{figure}[t]
    \centering
    \includegraphics[width=0.98\linewidth]{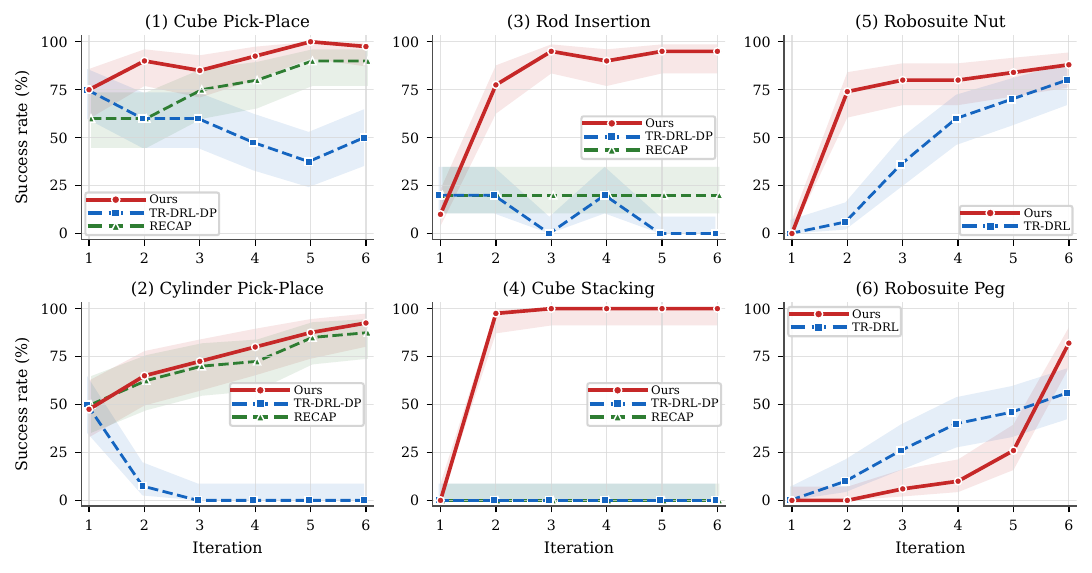}
    \caption{\wangcan{\textbf{Isaac Lab and Robosuite success curves.}
    We report four Isaac Lab tasks (1–4) and two Robosuite tasks (5–6).
    Each subplot reports Task~$\mathcal{H}$ success rate over self-improvement iterations.}}
    \label{fig:benchmark_success_curves}
\end{figure}

\begin{figure}[t]
    \centering
    \includegraphics[width=0.98\linewidth]{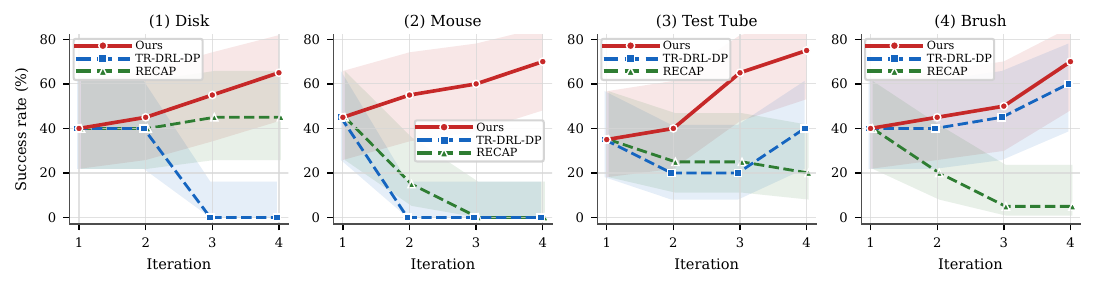}
    \caption{\textbf{Real-world robot success curves.}
    Each subplot reports Task~$\mathcal{H}$ success rate over self-improvement iterations.}
    \label{fig:realrobot_success_curves}
\end{figure}

\textbf{Overall performance.}
Across all three suites, Auto-E2H achieves monotonic gains over six self-improvement iterations on simulation and four iterations in the real world (Figures~\ref{fig:benchmark_success_curves} and~\ref{fig:realrobot_success_curves}).
Pooled across tasks within each suite, Task~$\mathcal{H}$ success rises from 33.1\% to 96.3\% on Isaac Lab (160 trials per iteration), from 0.0\% to 85.0\% on Robosuite (100 trials), and from 40.0\% to 70.0\% in the real world (80 trials), with non-overlapping pooled 95\% Wilson intervals between the initial and final policies in every suite.
The learned Task~$\mathcal{E}$ policy also serves as an automatic scene-reset module during real-world rollouts, succeeding in 80\%, 95\%, 82.5\%, and 80\% of cycles on Disk, Mouse, Test Tube, and Brush (84.4\% average), so manual intervention is needed only for the remaining failed reset cycles.
\wangcan{Detailed comparisons against TR-DRL-DP, RECAP, and Direct-$\mathcal{H}$ are reported in the corresponding subsections.}

\wangcan{\textbf{Isaac Lab benchmarks.}
In Figure~\ref{fig:benchmark_success_curves}, we compare Auto-E2H with TR-DRL-DP and RECAP.}
Over the six self-improvement iterations, the four-task Isaac Lab average rises monotonically from 33.1\% to 96.3\%, with per-task success ranging from 92.5\% to 100.0\% at the final iteration.
This exceeds \wangcan{TR-DRL-DP} and RECAP by 83.8 and 46.9 percentage points, respectively, and the gap is reflected in non-overlapping pooled 95\% Wilson intervals at the final iteration.

\textbf{Robosuite benchmarks.}
\wangcan{In Figure~\ref{fig:benchmark_success_curves}}, we also evaluate Auto-E2H on two Robosuite task pairs with reversible dynamics and asymmetric difficulty: nut assembly/disassembly and peg insertion/removal.
\wangcan{
We directly compare with TR-DRL rather than TR-DRL-DP, as the environments follow the released TR-DRL settings.
}
Starting from 0\% success at iteration 1, Auto-E2H rises to 85.0\% two-task average at iteration 6 (88.0\% nut, 82.0\% peg), exceeding TR-DRL by 17.0 percentage points.
\wangcan{Compared with this online RL technique,} the supervised policy-update signal of Auto-E2H produces stronger hard-task policies while remaining compatible with stable real-world rollouts.

\textbf{Real-world robot experiments.}
\wangcan{In Figure~\ref{fig:realrobot_success_curves},}
we apply the same iterative protocol to four real-world manipulation tasks: Disk, Mouse, Test Tube, and Brush.
Over four self-improvement iterations, the four-task average increases from 40.0\% (iteration 1) to 70.0\% (iteration 4), a 30.0 percentage-point gain over Auto-E2H's own initial policy trained on the 50 reverse demonstrations.
At the final iteration, Auto-E2H exceeds the TR-DRL-DP and RECAP four-task averages by 45.0 and 52.5 percentage points, respectively, with TR-DRL-DP and RECAP collapsing on three of the four real-world tasks.

\begin{figure}[t]
    \centering
    \includegraphics[width=0.98\linewidth]{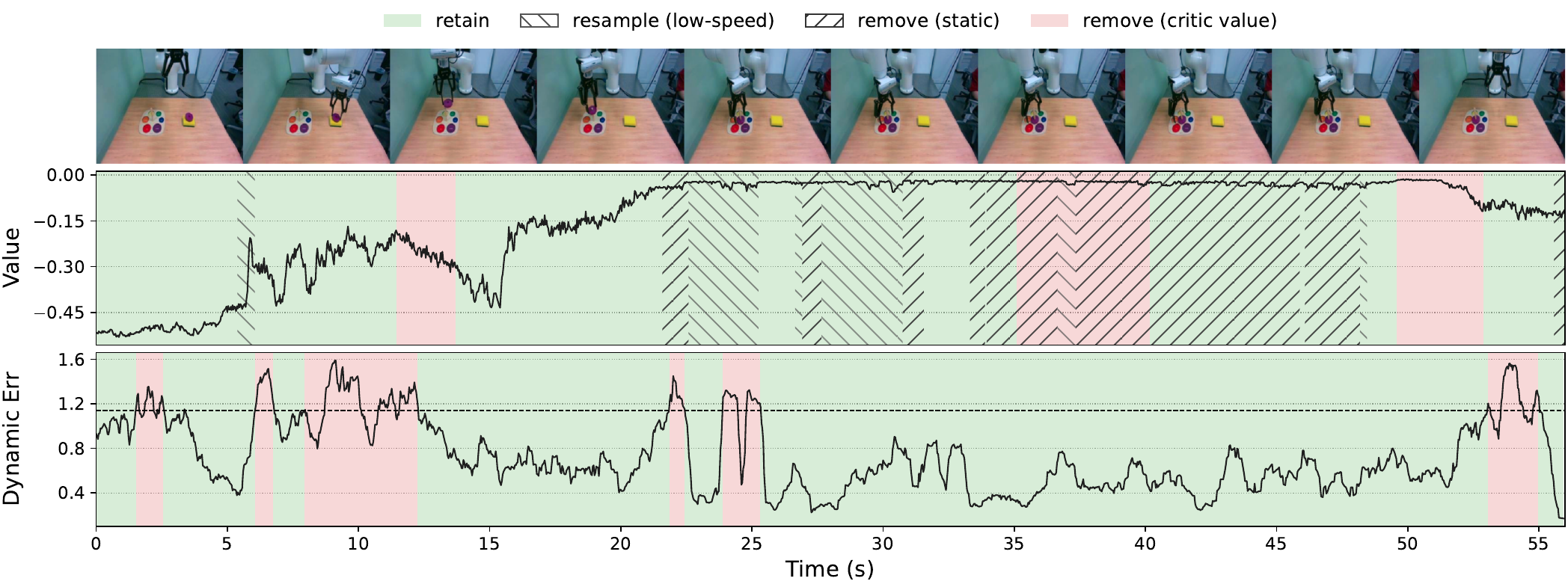}
    \caption{\textbf{Filtering comparison on a temporally reversed Disk environment rollout.}
    The top row shows uniformly sampled fixed-camera frames from the episode.
    The middle and bottom rows compare the filter scores used by our method and the baseline: green regions indicate retained timesteps, forward-hatched regions indicate static removal, backward-hatched regions indicate low-speed resampling, and red regions indicate critic- or dynamics-based removal.}
    \label{fig:filter_compare_realworld}
\end{figure}

\wangcan{\textbf{Filtering behavior comparison.}} Figure~\ref{fig:filter_compare_realworld} visualizes the filtering decisions on a representative temporally reversed rollout from the real-world task suite.
\wangcan{
Auto-E2H applies a hierarchical filtering process: static segments with minimal motion are removed, low-speed segments are resampled to regularize and accelerate motion, and segments with low critic value are pruned to exclude low-progress intervals. This allows Auto-E2H to adapt the retained data to Task~$\mathcal{H}$ behavior patterns and identify intervals where task progress is poor; for example, the critic value begins to drop around 12~s, corresponding to a misalignment between the wooden disk and the target rod that could lead to a skewed insertion failure.
}
By contrast, \wangcan{TR-DRL}'s dynamic filter mainly rejects segments with gripper-object contact or fast gripper motion that fall outside the model's prediction range, while retaining many static or hesitant frames with little environmental interaction; this mismatch explains its weaker improvement signal.

\subsection{Ablation Studies}
\begin{figure}[t]
    \centering
    \includegraphics[width=0.98\linewidth]{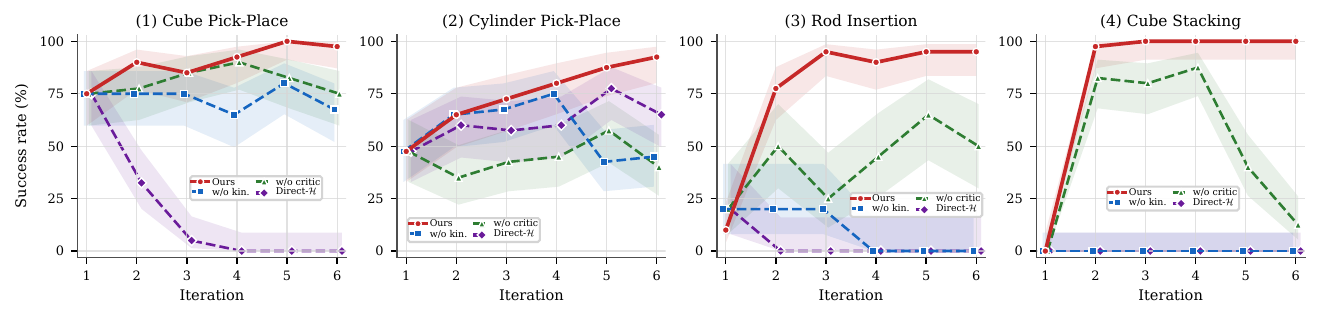}
    \caption{\textbf{Ablation success curves.} 
    \wangcan{We conduct ablation studies on the Isaac Lab environments.}
    Each subplot reports Task~$\mathcal{H}$ success rate over self-improvement iterations.}
    \label{fig:ablation_success_curves}
\end{figure}

\wangcan{We ablate Auto-E2H on four Isaac Lab tasks by removing the kinematic filter (w/o kin.), removing the critic-guided filter (w/o critic), or replacing reversed Task~$\mathcal{E}$ data with successful Task~$\mathcal{H}$ rollouts (Direct-$\mathcal{H}$).}
Figure~\ref{fig:ablation_success_curves} shows that the full pipeline is the most stable across tasks.
At the final iteration, the full Auto-E2H reaches 96.3\% average success, compared with 28.1\% without kinematic filtering, 44.4\% without critic-guided filtering, and 16.3\% with direct Task~$\mathcal{H}$ self-training.
The components are complementary: kinematic filtering removes static or low-speed artifacts, the critic selects progress-making reversed segments, and reverse-direction data avoids the sparsity of early hard-task successes.

\subsection{Data Collection Strategies}

\wangcan{Auto-E2H also reduces the human effort required to collect useful real-world data. We compare collecting 50 reversed easy-task demonstrations, used by our method, with directly collecting 50 hard-task demonstrations, using the same Diffusion Policy backbone and training setup as the iteration-1 Auto-E2H model. Table~\ref{tab:data_collection_cost} reports the average active recording time per demonstration, computed from the raw dataset captured at 30~Hz. The ``Ours easy demos'' column shows the success rate of Auto-E2H after the final iteration (Figure~\ref{fig:realrobot_success_curves}), while the ``Direct hard demos'' column corresponds to a policy trained from 50 hard-task demonstrations with the same model architecture and hyperparameters. Across all four tasks, Auto-E2H not only reduces collection time by 50–70\% per demonstration but also improves final task success by 10–20 percentage points.}


\begin{table}[t]
    \centering
    \small
    \setlength{\tabcolsep}{3.2pt}
    \caption{
    \wangcan{\textbf{Data collection strategy comparison.} Each task uses 50 demonstrations. Time reduction is $1-t_{\text{ours}}/t_{\text{direct}}$; success gain is in percentage points (pp).}}\label{tab:data_collection_cost}
    \begin{tabular}{@{}lcccccc@{}}
        \toprule
        & \multicolumn{2}{c}{Ours easy demos} & \multicolumn{2}{c}{Direct hard demos} & \multicolumn{2}{c}{Ours advantage} \\
        \cmidrule(lr){2-3}\cmidrule(lr){4-5}\cmidrule(l){6-7}
        Scene & Time/demo & Success & Time/demo & Success & Time reduction & Success gain \\
        \midrule
        Disk & \textbf{19.5 s} & \textbf{65\%} & 41.5 s & 50\% & \textbf{53.1\%} & \textbf{+15 pp} \\
        Mouse & \textbf{24.1 s} & \textbf{70\%} & 54.7 s & 60\% & \textbf{55.9\%} & \textbf{+10 pp} \\
        Test Tube & \textbf{19.6 s} & \textbf{75\%} & 39.1 s & 55\% & \textbf{50.0\%} & \textbf{+20 pp} \\
        Brush & \textbf{19.0 s} & \textbf{70\%} & 62.8 s & 60\% & \textbf{69.7\%} & \textbf{+10 pp} \\
        \bottomrule
    \end{tabular}
\end{table}



Across the four tasks, Auto-E2H reduces total operator time by about 60\% compared with hard-demo BC, including failed-reset recovery.

\section{Conclusion}
\label{sec:conclusion}

\wangcan{
We introduced Auto-E2H, a teleoperation-cost efficient framework for learning hard manipulation tasks by leveraging temporal reversal of easier tasks. By exploiting the inherent asymmetry between hard and easy tasks, Auto-E2H not only reduces dependence on costly expert demonstrations but also leverages reversed trajectories for rich supervision. To achieve this, the framework integrates three complementary components: closed-loop data collection, which autonomously generates diverse trajectories; hierarchical data refinement with kinematic and critic-guided filtering to retain high-quality segments; and iterative policy learning to progressively improve the hard-task policy. Experimental results across four Isaac Lab tasks, two Robosuite tasks, and four real-world manipulation tasks show that Auto-E2H consistently outperforms strong baselines. These results highlight that temporally reversed, easy-to-hard data can serve as scalable supervision, enabling efficient and reliable learning of complex manipulation policies with minimal human intervention.
}


\wangcan{\textbf{Limitations and Discussion. }Despite its advantages, Auto-E2H has limitations and is primarily suitable for directionally asymmetric tasks, where the reverse task is easier but still provides useful geometric supervision. It cannot fully handle general hard-task learning because reversed rollouts are inherently noisy rather than exact demonstrations, as contact, friction, gravity, and grasp/release events can break temporal reversibility. Pose-based reversal remains useful under moderate contact mismatch but degrades under strong direction-dependent resistance; force-aware or dynamics-aware control is needed beyond this regime. Dynamics-based reversal is further constrained because reliable world and inverse-dynamics models are difficult to learn in high-dimensional, contact-rich manipulation. Auto-E2H partially mitigates these issues through kinematic and critic-guided filtering, and closed-loop collection naturally diversifies easy-task rollouts via states induced by the evolving hard policy. Future work could extend its applicability by integrating stronger dynamics-based reversal, task-progress critics, and explicit diversity objectives for easy-task trajectories.
}

\clearpage
\section*{Acknowledgments and Disclosures}

This work is supported in part by the Research Grants Council (RGC) of the Hong Kong SAR under the General Research Fund (17203023), the Collaborative Research Fund (C5052-23G), and the NSFC/RGC Collaborative Research Scheme (CRS\_HKU703/24). The research work described in this paper was conducted in the JC STEM Lab of Multimedia and Machine Learning funded by The Hong Kong Jockey Club Charities Trust. We thank Selina Wang for discussions on the early idea of incremental data collection.

The authors declare no other competing interests relevant to this work.

\bibliographystyle{plainnat}
\bibliography{references}


\clearpage
\appendix
\section{Implementation Details}\label{app:implementation}

All learned policies use an image-conditioned Diffusion Policy~\cite{chi2023diffusion} with DDIM sampling~\cite{song2021denoising}.
The policy observes two history steps from a table camera, a wrist camera, and a 15-dimensional state containing end-effector pose, object pose, and gripper state.
In the real-robot experiments, the seven object-pose entries are fixed to zero during both training and inference. The policy uses only the two RGB images, the 7-D end-effector pose, and the 1-D gripper state; it does not require object-pose estimates or privileged object state.
The action is an 8-dimensional end-effector goal pose plus gripper command.
Data are recorded at $30\,\mathrm{Hz}$, and images are cropped to $128\times128$ during policy training.
The visual backbone uses a ResNet-18 image encoder.
The action horizon and executed action chunk both contain sixteen steps.
The diffusion model uses 100 training noise steps and ten DDIM denoising steps at inference.
The conditioning-dropout probability is 0.3.

Policy training uses Adam with learning rate $2\times10^{-4}$, seed 42, and batch size 256 per GPU\@.
The step count is scaled to preserve the same total sample budget under different GPU counts.
On the four NVIDIA A100 GPUs used in our experiments, each initial Task~$\mathcal{E}$ and Task~$\mathcal{H}$ policy is trained for 4000 optimizer steps, and each incremental Task~$\mathcal{H}$ update adds another 4000 optimizer steps.
At iteration $n$, the hard-task policy is warm-started from $\pi_h^{(n-1)}$ using $\mathcal{D}_h^{(0)}\cup\mathcal{D}_{\mathrm{filt}}^{(n)}$: the fixed initial reversed demonstrations and the current iteration's accepted data, sampled with a 1:1 weighting.
The Task~$\mathcal{E}$ policy is trained once and then kept fixed.

The critic is trained at each iteration from the current Task~$\mathcal{H}$ rollout data with Monte Carlo return labels and an 80/20 train/validation split.
It uses the same two-step visual/state observation format, a ResNet-18 image encoder initialized from the current action policy, an MLP with hidden dimensions $(512,512,256,256)$, MSE value loss, Adam learning rate $5\times10^{-4}$, weight decay $10^{-4}$, gradient clipping at 1.0, color jitter augmentation, dropout 0.15, crop size $112\times112$, batch size 256, and 500 training steps.

\paragraph{Filtering parameters.}
We use one fixed set of filtering parameters across all simulation and real-world environments.
The static filter uses end-effector speed threshold $\epsilon_s=0.04\,\mathrm{m/s}$, spatial-drift threshold $\epsilon_d=0.05$ m, and minimum static length $L_{\mathrm{s}}=16$ frames.
The speed filter uses low-speed threshold $\epsilon_l=0.06\,\mathrm{m/s}$ and resamples selected low-speed segments toward target step size $\Delta H\approx3.3\times10^{-3}$ m at 30 Hz, equivalent to a target speed of $0.1\,\mathrm{m/s}$.
For critic-guided filtering, we use $\gamma=1.0$, $\lambda=0$, smoothing window 31, set the advantage drop threshold $\alpha_{\mathrm{drop}}$ to the 30th percentile of smoothed advantage scores, and use minimum keep/drop length 30.

\paragraph{TR-DRL-DP.}
For the Isaac Lab comparison labeled TR-DRL-DP, we do not run the original SAC-based TR-DRL algorithm.
Instead, we keep the same closed-loop rollout collection and Diffusion Policy imitation-learning loop as our method, and replace our refinement-and-reversal module with a dynamics-model pipeline adapted from TR-DRL\@.
At each iteration, hard-task rollouts are encoded into visual-state features and used to train a forward dynamics model and an inverse dynamics model for 2000 steps each.
For every adjacent pair in an easy-task rollout, the inverse model predicts the reverse action from the later observation to the earlier observation, and the forward model checks that action by reconstructing the earlier feature/state.
The resulting cycle-consistency error is smoothed and thresholded at the 80th percentile for that iteration; transitions above the threshold are discarded, and retained runs shorter than 16 frames are removed.
The accepted easy-task segments are then temporally reversed.
This reversal is not implemented by simply reusing the recorded Task~$\mathcal{E}$ actions in reverse order: for a reversed step from original frame $t$ to frame $t-1$, the action label is predicted as $\mathrm{IDM}(o_t,o_{t-1})$, with the gripper dimension copied from the recorded gripper state.
The original RGB observations are kept, and the resulting hard-task-direction data are mixed with the original demonstrations to finetune the Diffusion Policy.
Thus this baseline isolates a TR-DRL-DP inside our imitation-learning setting, while the Robosuite TR-DRL comparison uses the original SAC-based implementation.

Policy and critic training were run on a four-GPU NVIDIA A100 server.
One Isaac Lab policy-update iteration, including rollout collection, filtering, critic training, and policy finetuning, typically took 1.5--3 hours depending on the task.

\FloatBarrier
\section{Evaluation Protocol and Statistical Reporting}\label{app:additional_results}

We evaluate Auto-E2H on four Isaac Lab task pairs, two Robosuite task pairs, and four real-world manipulation tasks.
The reported success curves are computed from success counts rather than smoothed training metrics.
The initial human-demonstration budgets are 40 easy-direction demonstrations for each Isaac Lab task pair, 10 demonstrations per Robosuite task pair, and 50 demonstrations per real-world task.
After initialization, no additional human demonstrations are collected.
Each self-improvement iteration instead adds policy rollouts: 40 cyclic Task~$\mathcal{H}$/Task~$\mathcal{E}$ rollout pairs per Isaac Lab task, 50 rollout episodes per Robosuite task, and 20 hard-task rollout episodes per real-world task.
Successful Task~$\mathcal{E}$ rollouts provide candidate reversed data, while all Task~$\mathcal{H}$ rollouts are used for critic training.
The rollout horizon is 3000 for cube pick-place, cylinder pick-place, and cube stacking, and 200 for rod insertion.

For the benchmark curves, each Isaac Lab task uses 40 evaluation trials, and each Robosuite task uses 50 trials.
For real-world experiments, each success rate is computed over 20 trials.
The shaded regions in the benchmark, real-world, and ablation success-curve figures show Wilson binomial confidence intervals~($z=1.96$), computed from the success-over-trial counts used by each plotted point.

\paragraph{Pooled suite-level confidence intervals.}
To complement the per-task curves, Table~\ref{tab:pooled_ci} reports pooled Task~$\mathcal{H}$ success for Auto-E2H at the initial and final self-improvement iterations of each suite, with 95\% Wilson binomial confidence intervals computed from the pooled successes-over-trials counts.
The initial- and final-iteration intervals do not overlap in any of the three suites.

\begin{table}[h]
    \centering
    \small
    \caption{Pooled Auto-E2H Task~$\mathcal{H}$ success and 95\% Wilson binomial confidence intervals at the initial and final self-improvement iterations.
    Here $k/n$ denotes successful trials over total evaluation trials; counts are summed across all tasks within each suite.}\label{tab:pooled_ci}
    \begin{tabular}{@{}lcccc@{}}
        \toprule
        Suite & Initial (k/n) & Initial 95\% CI & Final (k/n) & Final 95\% CI \\
        \midrule
        Isaac Lab (4 tasks)   & 53/160 (33.1\%) & [26.3\%, 40.7\%] & 154/160 (96.3\%) & [92.1\%, 98.3\%] \\
        Robosuite (2 tasks)   & 0/100 (0.0\%)   & [0.0\%, 3.7\%]   & 85/100 (85.0\%)  & [76.7\%, 90.7\%] \\
        Real-world (4 tasks)  & 32/80 (40.0\%)  & [30.0\%, 51.0\%] & 56/80 (70.0\%)   & [59.2\%, 78.9\%] \\
        \bottomrule
    \end{tabular}
\end{table}

\FloatBarrier
\section{Filter-Comparison Visualizations}\label{app:filter_visualizations}
Figure~\ref{fig:filter_compare_appendix_real_disk}--\ref{fig:filter_compare_appendix_robosuite_peg} provide per-scene examples of the filtering decisions used during easy-to-hard data conversion.
Each visualization shows sampled rollout frames, the critic value used by our filter, and the dynamics error used by the TR-DRL-DP.
Green regions indicate retained transitions, red regions indicate removed transitions, and the hatched regions mark segments removed as static or resampled due to low-speed motion.
The scenes are ordered as four real-world tasks, four Isaac Lab tasks, and two Robosuite tasks.
\begingroup
\raggedbottom
\captionsetup{skip=2pt,hypcap=false}
\noindent\begin{minipage}{\linewidth}
    \centering
    \includegraphics[width=0.98\linewidth]{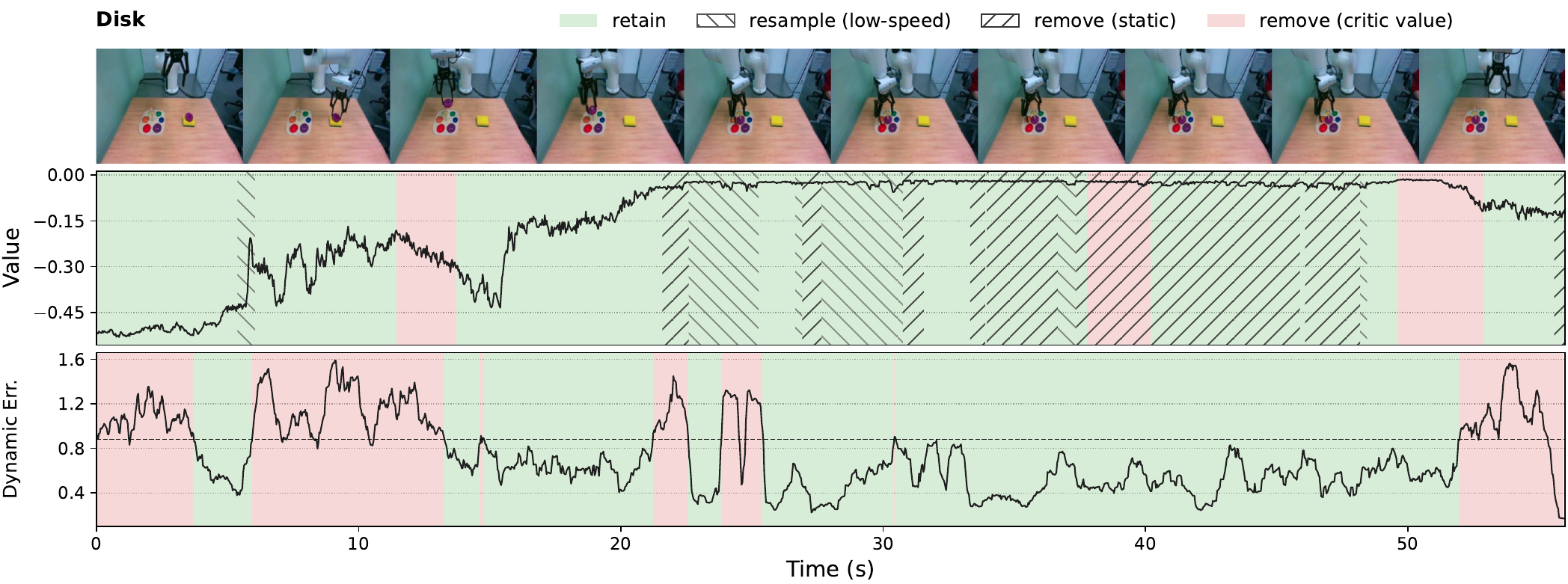}
    \captionof{figure}{Filter-comparison visualization for Real Disk.}
    \label{fig:filter_compare_appendix_real_disk}
\end{minipage}
\par\vspace{0.3em}

\noindent\begin{minipage}{\linewidth}
    \centering
    \includegraphics[width=0.98\linewidth]{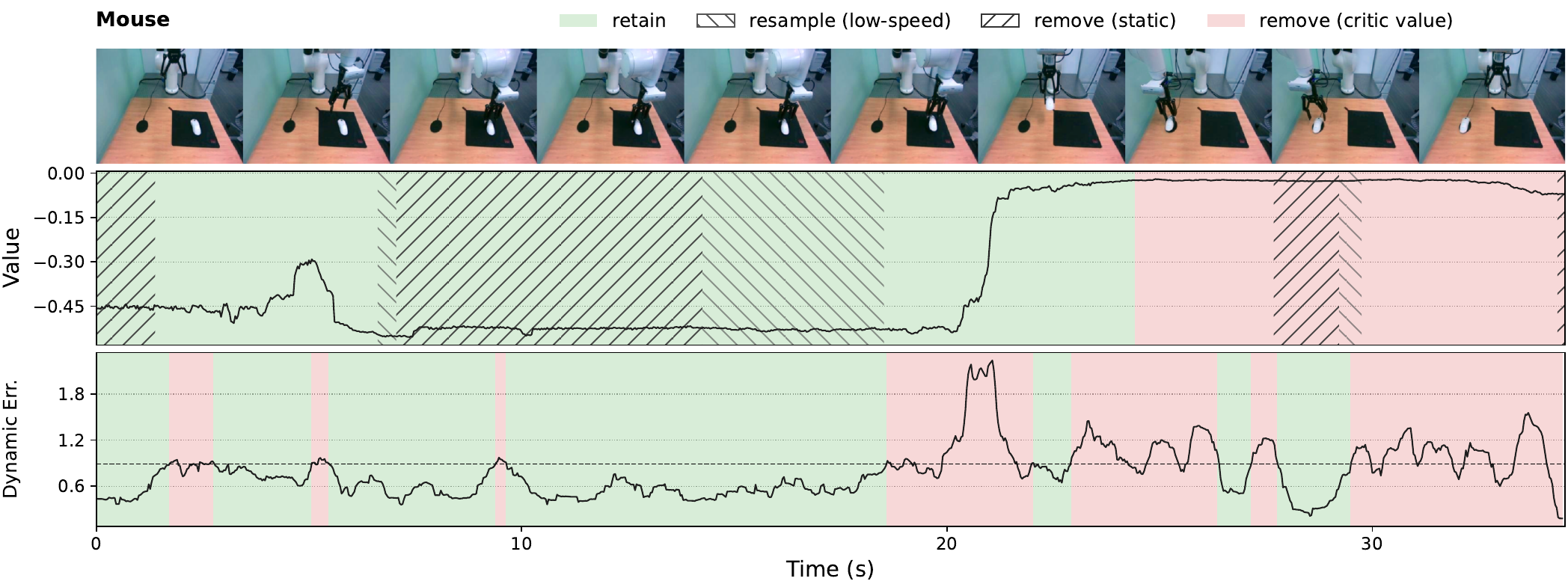}
    \captionof{figure}{Filter-comparison visualization for Real Mouse.}
    \label{fig:filter_compare_appendix_real_mouse}
\end{minipage}
\par\vspace{0.3em}

\noindent\begin{minipage}{\linewidth}
    \centering
    \includegraphics[width=0.98\linewidth]{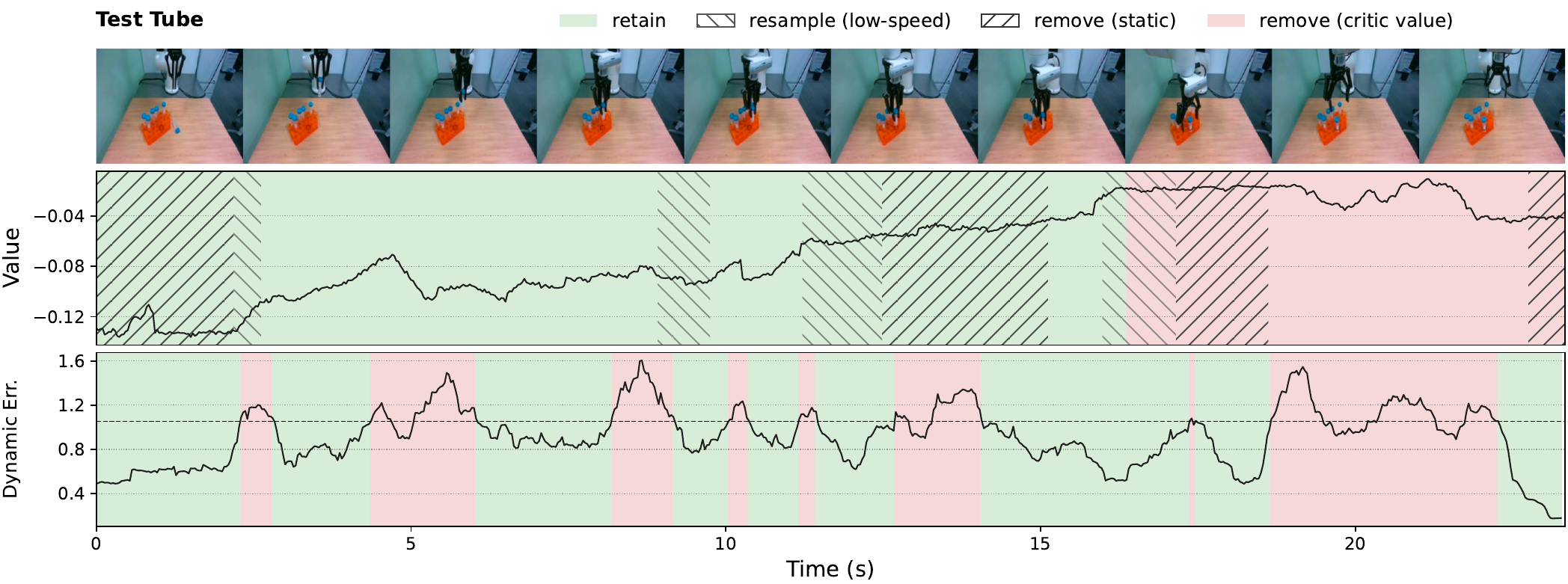}
    \captionof{figure}{Filter-comparison visualization for Real Test Tube.}
    \label{fig:filter_compare_appendix_real_test_tube}
\end{minipage}
\par\vspace{0.3em}

\noindent\begin{minipage}{\linewidth}
    \centering
    \includegraphics[width=0.98\linewidth]{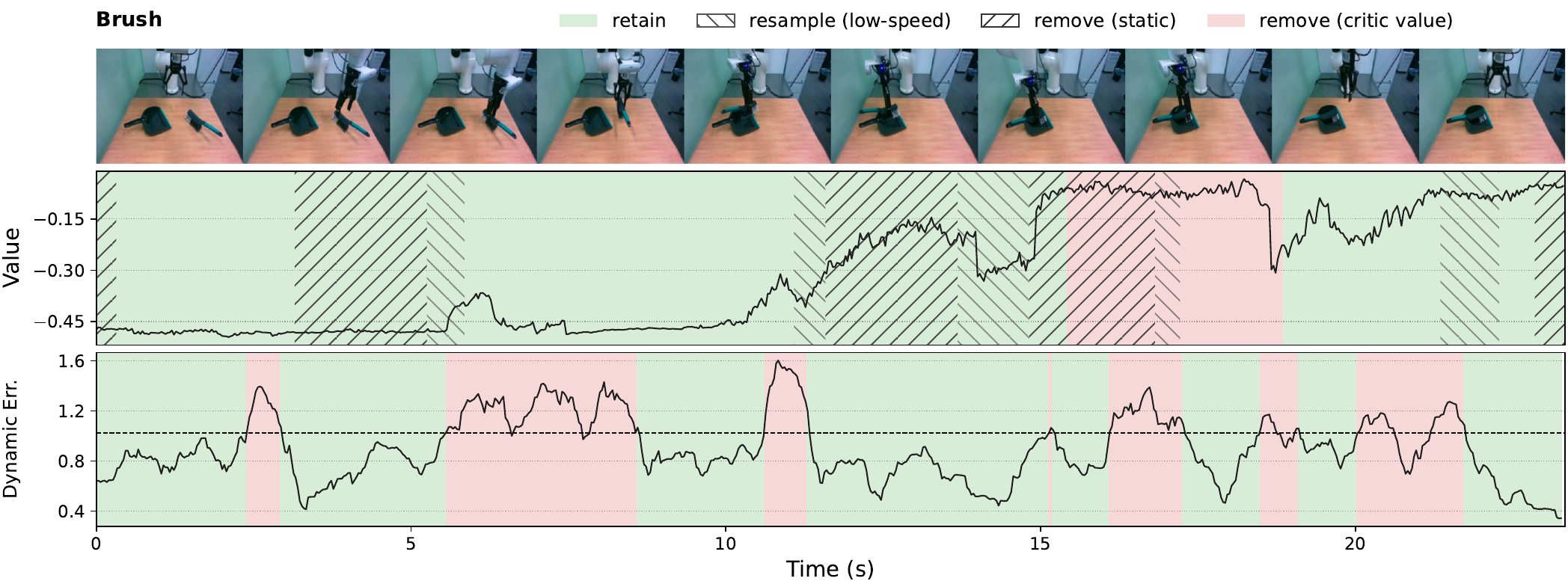}
    \captionof{figure}{Filter-comparison visualization for Real Brush.}
    \label{fig:filter_compare_appendix_real_brush}
\end{minipage}
\par\vspace{0.3em}

\noindent\begin{minipage}{\linewidth}
    \centering
    \includegraphics[width=0.98\linewidth]{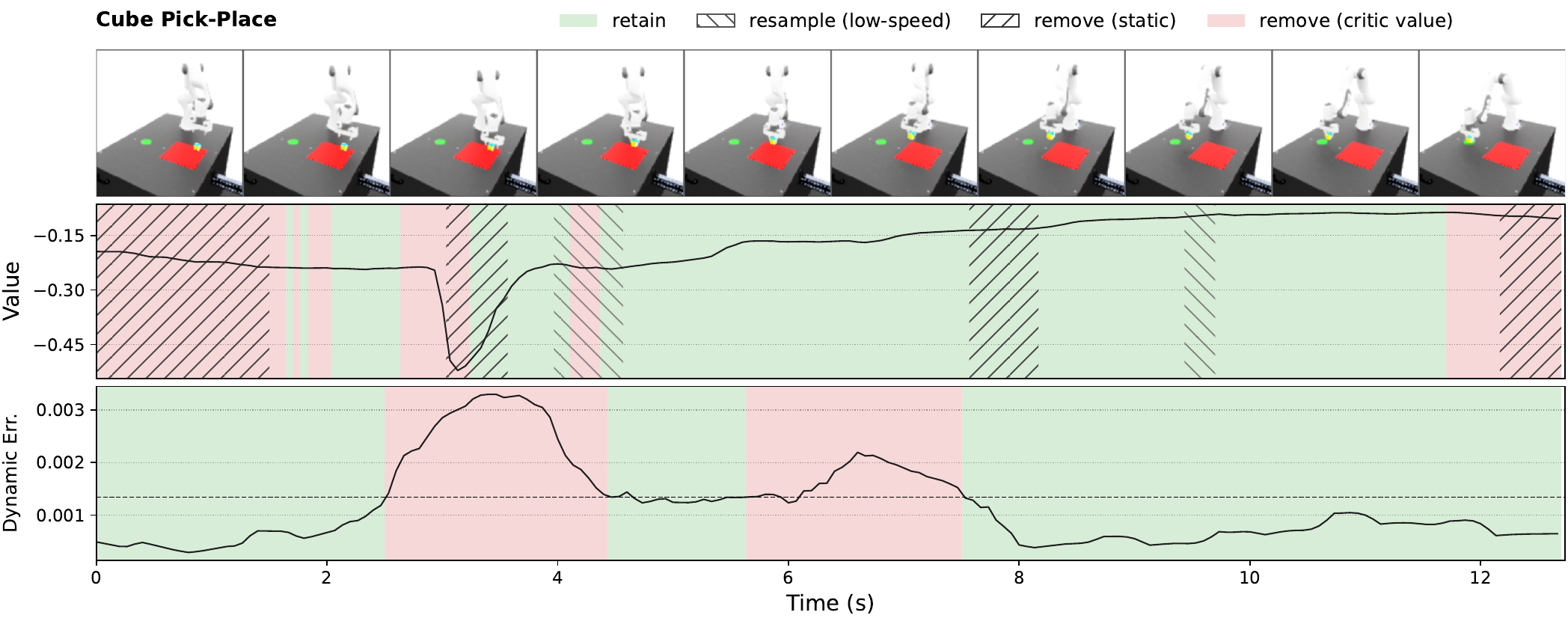}
    \captionof{figure}{Filter-comparison visualization for Isaac Cube Pick Place.}
    \label{fig:filter_compare_appendix_isaac_cube_pick_place}
\end{minipage}
\par\vspace{0.3em}

\noindent\begin{minipage}{\linewidth}
    \centering
    \includegraphics[width=0.98\linewidth]{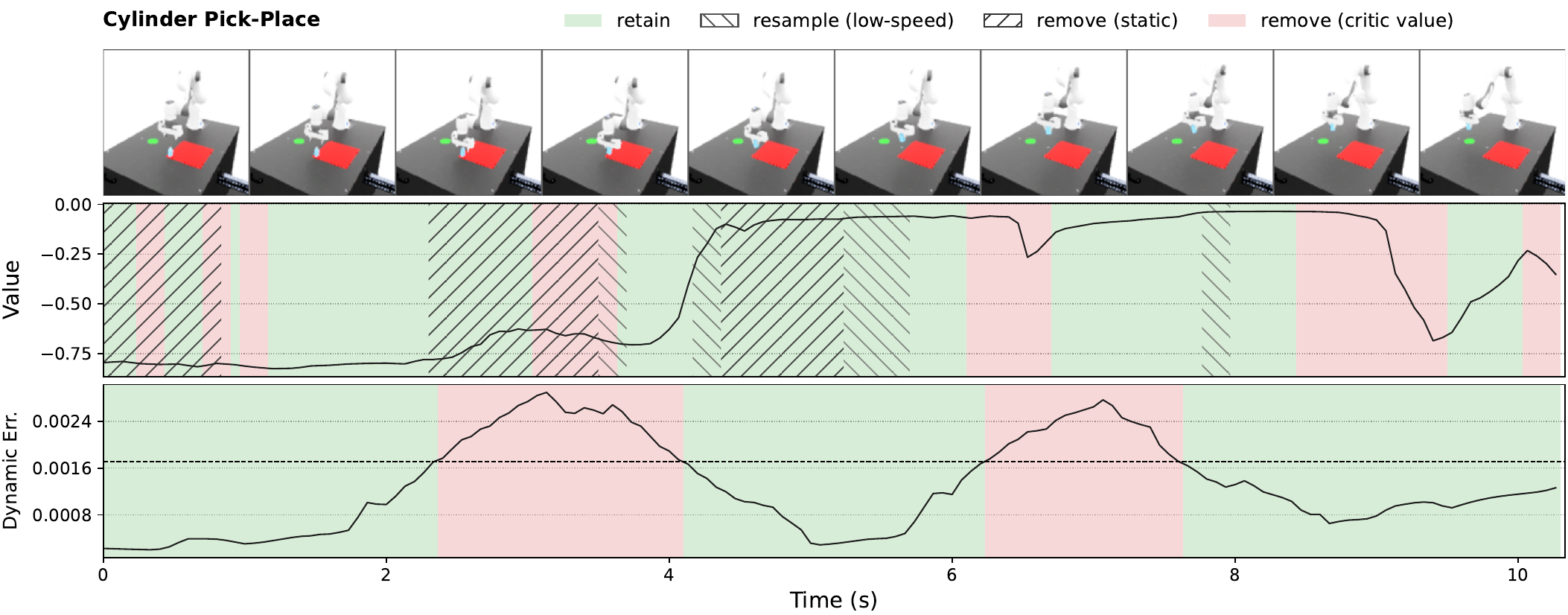}
    \captionof{figure}{Filter-comparison visualization for Isaac Cylinder Pick Place.}
    \label{fig:filter_compare_appendix_isaac_cylinder_pick_place}
\end{minipage}
\par\vspace{0.3em}

\noindent\begin{minipage}{\linewidth}
    \centering
    \includegraphics[width=0.98\linewidth]{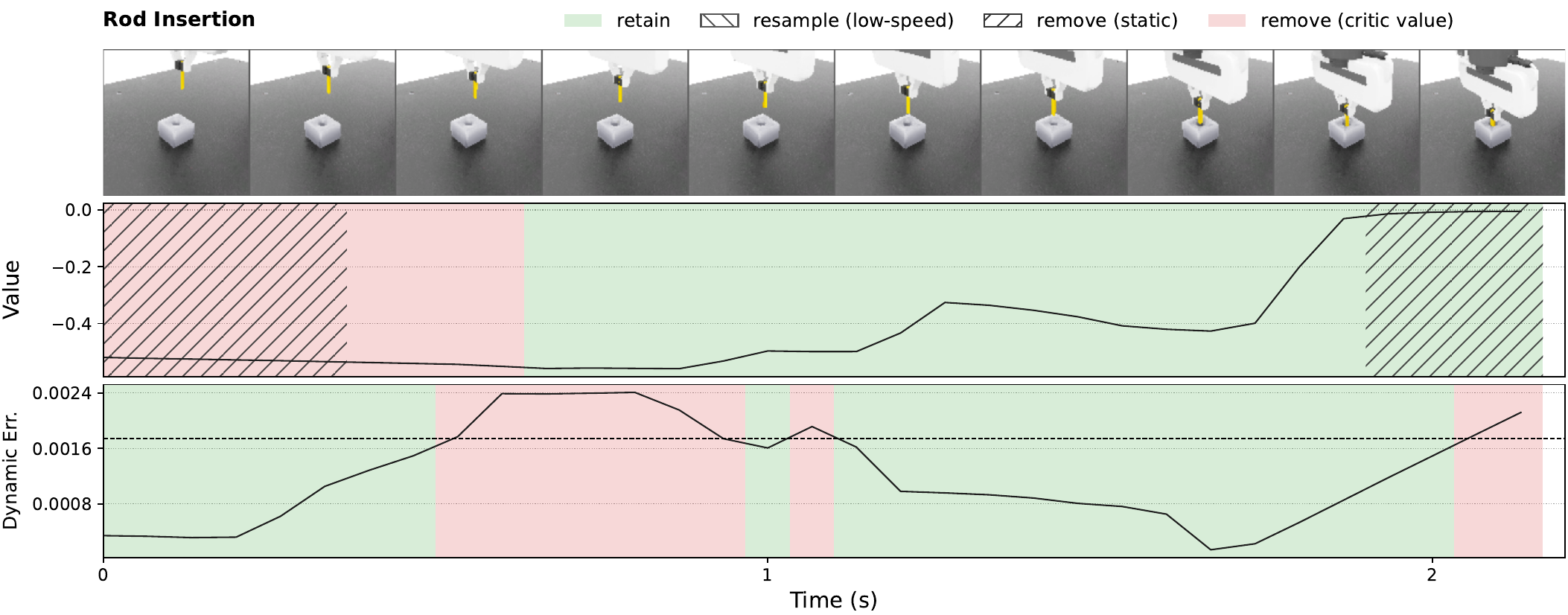}
    \captionof{figure}{Filter-comparison visualization for Isaac Rod Insertion.}
    \label{fig:filter_compare_appendix_isaac_rod_insertion}
\end{minipage}
\par\vspace{0.3em}

\noindent\begin{minipage}{\linewidth}
    \centering
    \includegraphics[width=0.98\linewidth]{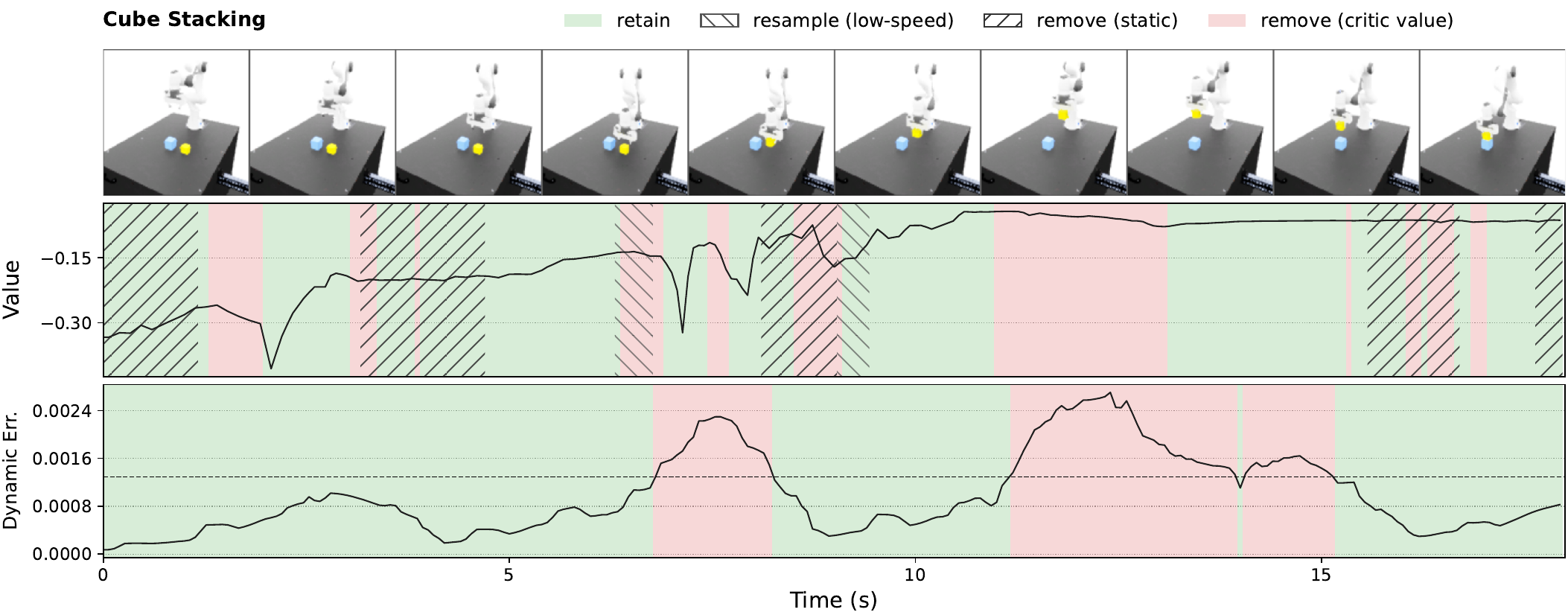}
    \captionof{figure}{Filter-comparison visualization for Isaac Cube Stack.}
    \label{fig:filter_compare_appendix_isaac_cube_stack}
\end{minipage}
\par\vspace{0.3em}

\noindent\begin{minipage}{\linewidth}
    \centering
    \includegraphics[width=0.98\linewidth]{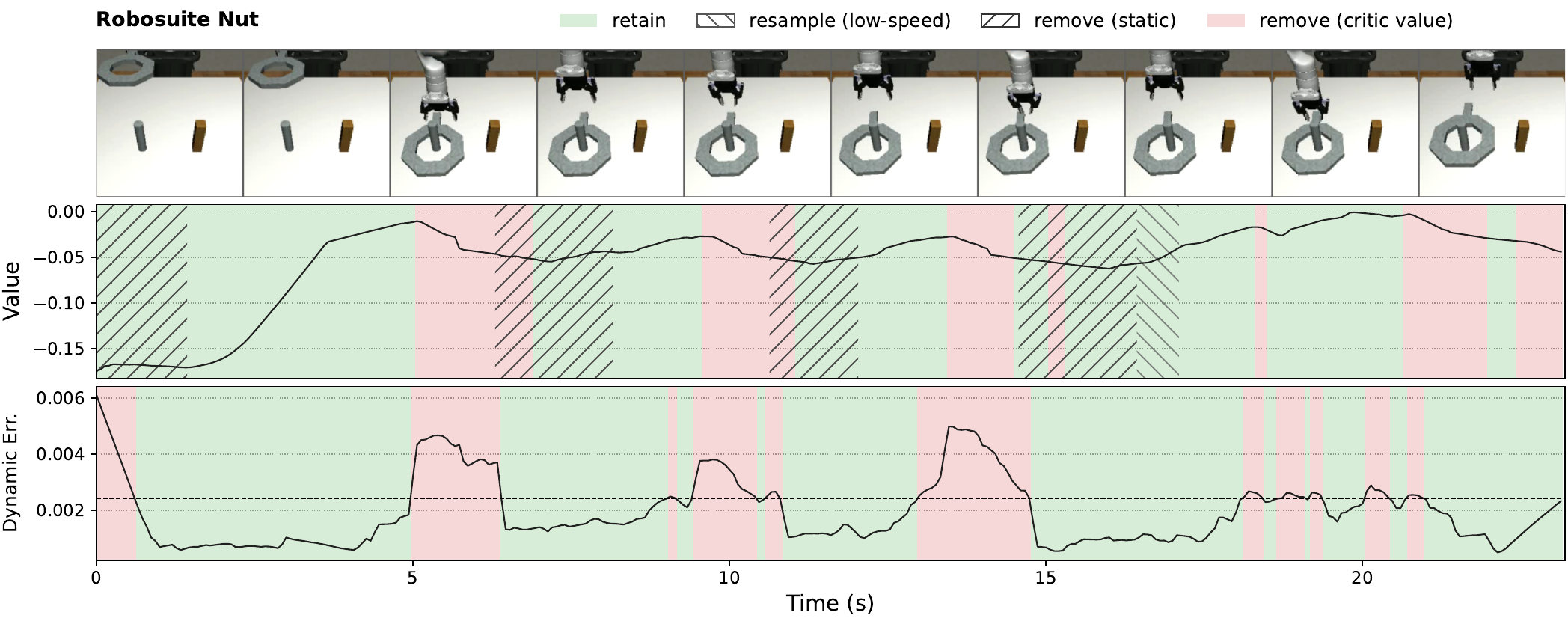}
    \captionof{figure}{Filter-comparison visualization for Robosuite Nut.}
    \label{fig:filter_compare_appendix_robosuite_nut}
\end{minipage}
\par\vspace{0.3em}

\noindent\begin{minipage}{\linewidth}
    \centering
    \includegraphics[width=0.98\linewidth]{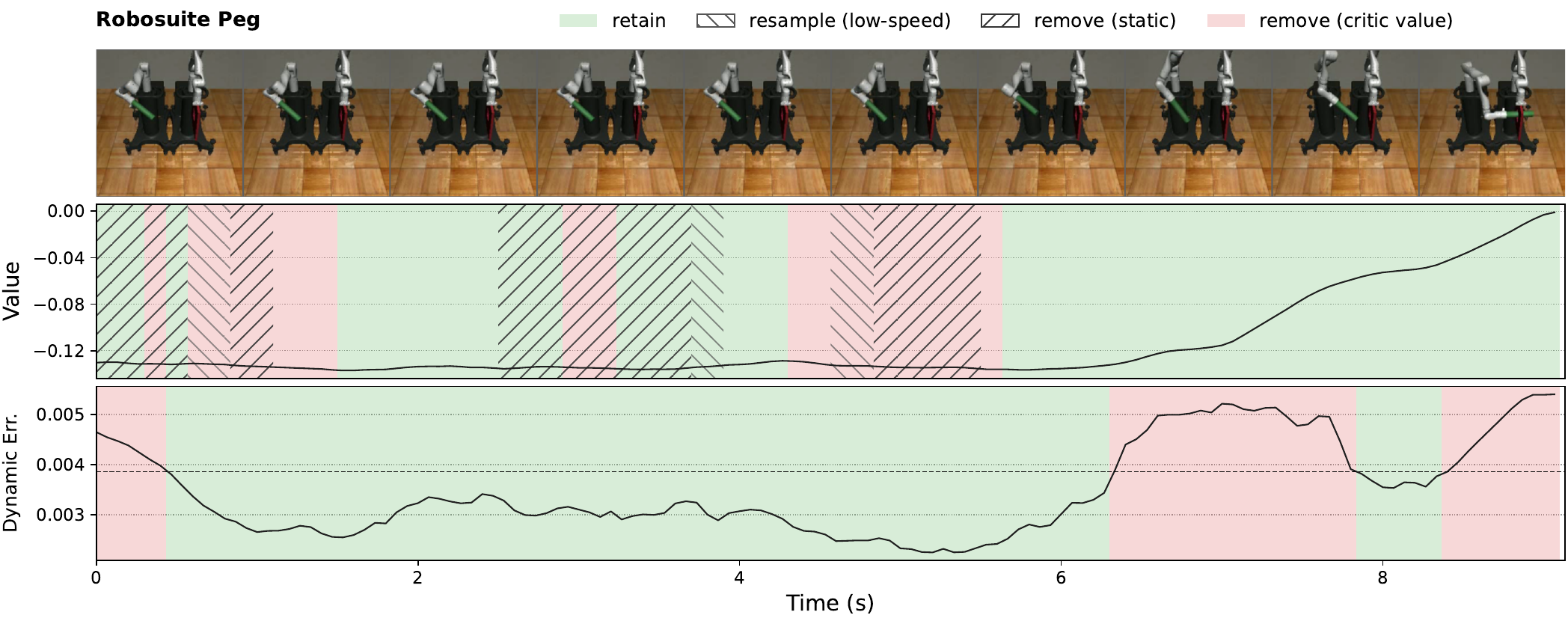}
    \captionof{figure}{Filter-comparison visualization for Robosuite Peg.}
    \label{fig:filter_compare_appendix_robosuite_peg}
\end{minipage}
\par\vspace{0.3em}

\flushbottom
\endgroup

\clearpage

\section{Critic Analysis}
\label{app:critic_analysis}

We study critic filtering and update choices on four Isaac Lab tasks: cube pick-place, cylinder pick-place, rod insertion, and cube stacking.
The comparisons use the same Diffusion Policy backbone and training budget, six policy iterations, and 40 evaluation trials per reported success rate.

\paragraph{Cold-start supervision.}
For success indicator $y\in\{0,1\}$, rollout length $T\leq H$, horizon $H$, and frame index $t$, the critic target is $\bar{G}_t=-(T-t)/(2H)-(1-y)/2$.
It lies in $[-0.5,0)$ for successful rollouts and $[-1,-0.5)$ for failed rollouts.
If every rollout fails at the maximum horizon, neighboring frames still differ by $1/(2H)$ in their targets.
This provides temporal-ordering supervision, but does not by itself establish that the critic is a calibrated success predictor or a reliable task-progress measure in every state.
Action reconstruction and kinematic filtering also contribute to cold start before critic filtering is applied.

\paragraph{Early filtering and critic updates.}
Table~\ref{tab:critic_ablation_extra} reports two controlled comparisons on the four Isaac Lab tasks: bypassing critic filtering only in iterations 1--2, and fine-tuning the critic from the previous iteration rather than retraining it.
The early-filter comparison keeps reversal, kinematic filtering, training budget, evaluation starts, and seed fixed.
Removing the early critic filter reduces iteration-2 success from 77.5\% to 50\% on rod insertion and from 100\% to 82.5\% on stacking.
After filtering is restored, the final differences are only 0--2.5 percentage points across tasks.
Retraining is generally slightly better in intermediate iterations, while final results are comparable; we retain retraining as the default.

\begin{table}[htbp]
\centering
\small
\caption{Critic-filter and update ablations. Entries are hard-task success rates (\%), each over 40 evaluation trials. ``No early filter'' bypasses critic filtering only in iterations 1--2; ``Warm-start critic'' fine-tunes the preceding critic.}
\label{tab:critic_ablation_extra}
\begin{tabular}{@{}llrrrrrr@{}}
\toprule
Task & Variant & Iter. 1 & 2 & 3 & 4 & 5 & 6 \\
\midrule
Cube pick-place & Full method & 75 & 95 & 85 & 92.5 & 100 & 97.5 \\
& No early filter & 75 & 82.5 & 85 & 92.5 & 100 & 100 \\
& Warm-start critic & 77.5 & 82.5 & 87.5 & 90 & 95 & 100 \\
\addlinespace
Cylinder pick-place & Full method & 45 & 65 & 75 & 80 & 80 & 85 \\
& No early filter & 40 & 65 & 70 & 77.5 & 80 & 82.5 \\
& Warm-start critic & 42.5 & 65 & 72.5 & 75 & 82.5 & 82.5 \\
\addlinespace
Rod insertion & Full method & 10 & 77.5 & 95 & 90 & 95 & 95 \\
& No early filter & 10 & 50 & 87.5 & 90 & 95 & 95 \\
& Warm-start critic & 10 & 67.5 & 90 & 90 & 92.5 & 95 \\
\addlinespace
Cube stacking & Full method & 0 & 100 & 100 & 100 & 100 & 100 \\
& No early filter & 0 & 82.5 & 100 & 100 & 100 & 100 \\
& Warm-start critic & 0 & 92.5 & 100 & 100 & 100 & 100 \\
\bottomrule
\end{tabular}
\end{table}

\FloatBarrier
\section{Training-Seed Variability}
\label{app:seed_analysis}

We measure the training variability of Auto-E2H on cube pick-place, cylinder pick-place, rod insertion, and cube stacking in Isaac Lab.
Table~\ref{tab:training_seeds_extra} reports three independent training seeds (42, 0, and 1), with 40 evaluation trials per entry, across six policy iterations.
Final success is 97.5--100\% for cube pick-place, 82.5--92.5\% for cylinder pick-place, 95--97.5\% for rod insertion, and 100\% for cube stacking.
These runs are reported separately from the original benchmark curves.
Differences across training seeds measure training variability; a Wilson interval for one policy's evaluation trials instead describes binomial evaluation uncertainty, rather than variability across training seeds.

\begin{table}[htbp]
\centering
\small
\caption{Training-seed variability on Isaac Lab. Entries are hard-task success rates (\%), with 40 evaluation trials each.}
\label{tab:training_seeds_extra}
\begin{tabular}{@{}llrrrrrr@{}}
\toprule
Task & Training seed & Iter. 1 & 2 & 3 & 4 & 5 & 6 \\
\midrule
Cube pick-place & 42 & 75 & 95 & 85 & 92.5 & 100 & 97.5 \\
& 0 & 62.5 & 75 & 87.5 & 92.5 & 100 & 100 \\
& 1 & 67.5 & 85 & 87.5 & 95 & 95 & 100 \\
\addlinespace
Cylinder pick-place & 42 & 45 & 65 & 75 & 80 & 80 & 85 \\
& 0 & 40 & 62.5 & 65 & 75 & 80 & 82.5 \\
& 1 & 50 & 70 & 77.5 & 77.5 & 85 & 92.5 \\
\addlinespace
Rod insertion & 42 & 10 & 77.5 & 95 & 90 & 95 & 95 \\
& 0 & 12.5 & 80 & 82.5 & 90 & 95 & 95 \\
& 1 & 2.5 & 70 & 87.5 & 90 & 92.5 & 97.5 \\
\addlinespace
Cube stacking & 42 & 0 & 100 & 100 & 100 & 100 & 100 \\
& 0 & 2.5 & 97.5 & 100 & 100 & 100 & 100 \\
& 1 & 2.5 & 100 & 100 & 100 & 100 & 100 \\
\bottomrule
\end{tabular}
\end{table}

\FloatBarrier
\section{Data Collection and Training Costs}
\label{app:cost_audit}

Each Auto-E2H run uses 50 easy-task demonstrations and 60 autonomous collection episodes; hard-demo BC uses 50 hard-task demonstrations.
Failed resets require 21--45~s of operator recovery per task over the full run.
Including recovery, Auto-E2H reduces total operator time by about 60\% across the four real-world tasks (Table~\ref{tab:operator_audit}).

\begin{table}[htbp]
\centering
\small
\caption{Per-task time costs in seconds. Operator time includes demonstration collection and failed-reset recovery. Training time sums initial policy training, critic training, and policy updates. Autonomous rollout time is for Auto-E2H.}
\label{tab:operator_audit}
\begin{tabular}{@{}lrrrrr@{}}
\toprule
& \multicolumn{2}{c}{Operator time} & Auto rollout & \multicolumn{2}{c}{Training time} \\
\cmidrule(lr){2-3}\cmidrule(l){5-6}
Task & Auto-E2H & Hard-demo BC & Auto-E2H & Auto-E2H & Hard-demo BC \\
\midrule
Disk & 997 & 2077 & 2472 & 39151 & 35048 \\
Mouse & 1251 & 2734 & 1909 & 40328 & 37926 \\
Test Tube & 1022 & 1957 & 1365 & 39104 & 35109 \\
Brush & 971 & 3141 & 1030 & 36176 & 39148 \\
\bottomrule
\end{tabular}
\end{table}

\FloatBarrier
\section{Boundary of Pose-Based Reversal under Contact Resistance}
\label{app:dynamics_boundary}

Auto-E2H reconstructs motion geometry rather than interaction dynamics.
Contact wrenches depend on friction, compliance, contact mode, and loading direction, and generally cannot be recovered by reversing their order or sign.
The action space contains no force command and the policy receives no force-torque input.
The following controlled experiments quantify the applicability boundary in the tested tasks; their force values are not universal thresholds for other assemblies or controllers.

\subsection{Controlled Simulation}
We fix geometry, demonstrations, and controller in Isaac Sim rod insertion and apply a simulated friction force opposite to instantaneous rod motion, with direction updated as the rod moves rather than constrained to its axis.
Reconstructed actions are first replayed on 100 paired initial states (seed 42), without policy training.
Extraction succeeds throughout, while reversed insertion declines sharply above 2~N (Table~\ref{tab:resistance_replay}).
Extraction produces a tension-like load that tends to preserve alignment, whereas insertion produces a compression-like load that can amplify angular error and rod--gripper slip.
Every replay failure above 2~N involves more than 5~mm of slip.
The complete learning pipeline follows the same trend (Table~\ref{tab:resistance_learning}).
At up to 2~N, iteration-6 success reaches 92.5--95\%; at 4 and 6~N it falls to 17.5\% and 0\%.

\begin{table}[htbp]
\centering
\small
\caption{Direct replay under simulated contact resistance, without policy training. Entries are successful trials out of 100 paired initial states.}
\label{tab:resistance_replay}
\begin{tabular}{@{}rrr@{}}
\toprule
Simulated friction force (N) & Extraction & Reversed replay \\
\midrule
0 & 100/100 & 100/100 \\
0.5 & 100/100 & 100/100 \\
1 & 100/100 & 100/100 \\
2 & 100/100 & 97/100 \\
4 & 100/100 & 21/100 \\
6 & 100/100 & 17/100 \\
\bottomrule
\end{tabular}
\end{table}

\begin{table}[htbp]
\centering
\small
\caption{Complete Auto-E2H learning under simulated resistance. Entries are hard-task success rates (\%), each over 40 evaluation trials.}
\label{tab:resistance_learning}
\begin{tabular}{@{}lrrrrrr@{}}
\toprule
Resistance (N) & Iter. 1 & 2 & 3 & 4 & 5 & 6 \\
\midrule
0 & 10 & 77.5 & 95 & 90 & 95 & 95 \\
0.5 & 5 & 67.5 & 90 & 90 & 92.5 & 95 \\
1 & 7.5 & 72.5 & 85 & 87.5 & 92.5 & 92.5 \\
2 & 5 & 65 & 87.5 & 90 & 87.5 & 92.5 \\
4 & 0 & 5 & 12.5 & 15 & 12.5 & 17.5 \\
6 & 0 & 0 & 0 & 0 & 2.5 & 0 \\
\bottomrule
\end{tabular}
\end{table}

\FloatBarrier
\subsection{Real-Robot Disk Insertion}
The hard direction grasps a disk and inserts it over a vertical rod.
The untaped rod diameter is 13.91~mm and the disk-hole diameter is 16.14~mm.
We progressively increase contact resistance by wrapping the rod in 0.1-mm-thick paper masking tape; each full wrap increases diameter by approximately 0.2~mm, with realized diameters measured after wrapping.
The wrist force-torque sensor is used for analysis, not as a policy input.
For successful insertions, the peak-force statistic is the mean of the 30 largest force magnitudes among frames with nonzero vertical motion.
It is an empirical proxy for contact resistance, not an estimate of the friction coefficient.

Table~\ref{tab:real_force_boundary} reports success over 20 evaluation trials per entry.
At 0--6 wraps (3.4--3.6~N), reversed data remains useful, reaching 60--70\% final success.
At 7 wraps (10.1~N), performance degrades substantially, although additional reversed data improves success from 20\% to 30\%.
At 8 wraps no insertion succeeds and the robot triggers its safety stop.
The load then reflects compression from a jammed fit rather than sliding resistance, and no successful trajectory exists from which to compute the statistic, so it is reported as N/A.
These results motivate force-aware or dynamics-aware extensions for strongly direction-dependent contact.

\begin{table}[htbp]
\centering
\small
\caption{Real-robot contact boundary. The peak-force statistic is computed only from successful insertions. Iteration columns report success rates (\%), each over 20 trials.}
\label{tab:real_force_boundary}
\begin{tabular}{@{}rrrrrrr@{}}
\toprule
Peak force (N) & Tape wraps & Rod diameter (mm) & Iter. 1 & 2 & 3 & 4 \\
\midrule
3.4 & 0 & 13.91 & 40 & 45 & 55 & 65 \\
3.5 & 3 & 14.51 & 40 & 45 & 60 & 70 \\
3.6 & 6 & 15.09 & 35 & 45 & 55 & 60 \\
10.1 & 7 & 15.46 & 20 & 20 & 25 & 30 \\
N/A & 8 & 15.78 & 0 & 0 & 0 & 0 \\
\bottomrule
\end{tabular}
\end{table}

\clearpage
\section{Multi-Robot Operator-Load Estimate}
\label{app:operator_scaling}

We estimate a possible concurrent-deployment workload from the measured single-robot audit; this is not a multi-robot deployment experiment.
Assume each of $N$ robots reproduces the single-robot rollout time $T_{\mathrm{auto}}$ and intervention time $T_{\mathrm{int}}$, and all robots collect concurrently, without credit for staggered scheduling or operator-free fine-tuning intervals.
The measured $(T_{\mathrm{auto}},T_{\mathrm{int}})$ pairs, in seconds, are $(2472,21)$ for Disk, $(1909,45)$ for Mouse, $(1365,42)$ for Test Tube, and $(1030,23)$ for Brush.
The implied average utilization of one shared operator is
\begin{equation}
\rho_N = N\frac{T_{\mathrm{int}}}{T_{\mathrm{auto}}}.
\end{equation}
For ten robots, this gives 8.5\% for Disk, 23.6\% for Mouse, 30.8\% for Test Tube, and 22.3\% for Brush.
These averages do not bound simultaneous alerts or waiting times.
A proposed implementation can stagger rollout and fine-tuning cycles and queue remaining alerts, pausing only the robot awaiting recovery while the others continue collecting.
This scheduling proposal may reduce peak contention, but its benefit is not measured by the single-robot experiment.

\end{document}